\PassOptionsToPackage{table}{xcolor}
\documentclass[11pt]{article}
\usepackage{acl}

\definecolor{tableheader}{RGB}{235,241,248}
\definecolor{tableoverall}{RGB}{218,231,245}
\definecolor{tablebest}{RGB}{205,222,243}

\usepackage[T1]{fontenc}
\usepackage[utf8]{inputenc}
\usepackage{times}
\usepackage{latexsym}
\usepackage{microtype}
\usepackage{inconsolata}

\usepackage{amsmath}
\usepackage{amssymb}
\usepackage{dsfont}
\usepackage{pifont}

\usepackage{booktabs}
\usepackage{multirow}
\usepackage{tabularx}
\usepackage{array}

\usepackage{graphicx}
\usepackage{tikz}

\usepackage{dblfloatfix}

\usepackage{enumitem}
\usepackage{setspace}
\usepackage[most]{tcolorbox}

\usepackage{algorithm}
\usepackage{algpseudocode}
\usepackage{caption}
\usepackage{listings}
\lstdefinestyle{sqls}{
    language=SQL,
    basicstyle=\ttfamily,
    keywordstyle=\ttfamily,
    stringstyle=\ttfamily,
    commentstyle=\ttfamily\itshape,
    showstringspaces=false,
    breaklines=true,
    frame=none,
    numbers=none,
    escapeinside={(*@}{@*)},
    morekeywords={SELECT, FROM, WHERE, JOIN, ON, GROUP, BY, ORDER, HAVING,
        DISTINCT, AS, AND, OR, NOT, NULL, IS, IN, LIKE, BETWEEN, LIMIT, OFFSET,
        INNER, LEFT, RIGHT, OUTER, UNION, INTERSECT, EXCEPT, COUNT, SUM, AVG,
        MAX, MIN, ROUND, CAST}
}
\newcommand{\sql}[1]{\lstinline[style=sqls]|#1|}

\newcommand{\namedpara}[1]{%
  \par\vspace{1.4ex}%
  \noindent\textbf{#1}\hspace{0.35em}\ignorespaces
}

\newcommand{\casehead}[1]{{\small\bfseries[#1]}}

\newcommand{\sqlbox}[2]{%
  \begingroup
  \setlength{\fboxsep}{3pt}%
  \setlength{\fboxrule}{0.25pt}%
  \noindent\fcolorbox{black!15}{black!4}{%
    \parbox{\dimexpr\linewidth-2\fboxsep-2\fboxrule\relax}{%
      \raggedright
      \scriptsize
      \textbf{\color{black!82}#1}\enspace
      {\ttfamily\color{black!70}#2}%
    }%
  }%
  \endgroup
}

\usepackage{lipsum}

\title{Evaluating LLMs on Conversational Text-to-SQL under \\ Chain Ambiguity and Intent Drift}

\author{
Yujia Liu\textsuperscript{1*}, 
Jiayan Lin\textsuperscript{1*}, 
Zijin Hong\textsuperscript{2*\dag}, 
Zheng Yuan\textsuperscript{2}, 
Shengyuan Chen\textsuperscript{2}, 
Hao Chen\textsuperscript{3}\\
{\bf Qinggang Zhang\textsuperscript{4}, Xiao Huang\textsuperscript{2}, Feiran Huang\textsuperscript{5}\footnotemark[2]}\\ 
\textsuperscript{1}Jinan University \textsuperscript{2}The Hong Kong Polytechnic University \\
\textsuperscript{3}City University of Macau
\textsuperscript{4}Jilin University
\textsuperscript{5}Beihang University
\\ 
\texttt{yujialiu@stu2023.jnu.edu.cn}; \texttt{ljiayan@stu2023.jnu.edu.cn}\\ \texttt{zijin.hong@connect.polyu.hk}; 
\texttt{xiao.huang@polyu.edu.hk};
\texttt{huangfr@buaa.edu.cn}
}

\begin{document}
\maketitle
\renewcommand{\thefootnote}{\fnsymbol{footnote}}
\footnotetext[1]{All authors contributed equally to this research.}
\footnotetext[2]{Corresponding author.}
\renewcommand{\thefootnote}{\arabic{footnote}}

\begin{abstract}
\vspace{-3mm}
Recent advances in large language models (LLMs) have established conversational text-to-SQL as a practical interface between users and databases, often involving multiple turns of clarification and revision. However, existing benchmarks primarily evaluate execution accuracy, leaving the unfolding and shifting of user intent across turns largely uncovered. To address this, we introduce \textbf{\textsc{TIDE-Bench}}, a benchmark for conversa\textbf{\textsc{ti}}onal text-to-SQL under chain ambiguity and intent \textbf{\textsc{d}}rift \textbf{\textsc{e}}valuation, targeting two recurring patterns: \textbf{chain ambiguity}, where an underspecified question triggers layered clarification with conditional dependencies, and \textbf{intent drift}, where the user retracts and replaces a previously committed request element. Built on 514 anchor SQLs from BIRD, \textsc{TIDE-Bench} comprises 1{,}542 samples and introduces dedicated metrics for chain identification and drift recognition-resolution beyond execution accuracy. Evaluating 12 advanced LLMs reveals a persistent chain identification bottleneck unaffected by clarification frequency, a wide drift recognition-resolution gap, and overlap between failure modes when jointly activated.
The corresponding code of \textsc{TIDE-Bench} is released for further research\footnote{\href{https://github.com/Rcrossmeister/TIDE-Bench}{https://github.com/Rcrossmeister/TIDE-Bench}}. 
\end{abstract}

\section{Introduction}

The advent of large language models (LLMs) has established text-to-SQL~\cite{hong2025next,liu2025survey} as a practical interface between users and structured databases.
In real deployments, this interaction is rarely single-turn: users and agents converse over multiple rounds, refining requests and seeking clarification before SQL is produced~\cite{yu2019cosql,huo2026interact}.
Even agents that perform well in single-turn evaluation can fail once conversation becomes part of the task ~\cite{yu2018spider,li2023bird,lei2025spider2}.

\begin{figure}[!t]
    \centering
    \includegraphics[width=\linewidth]{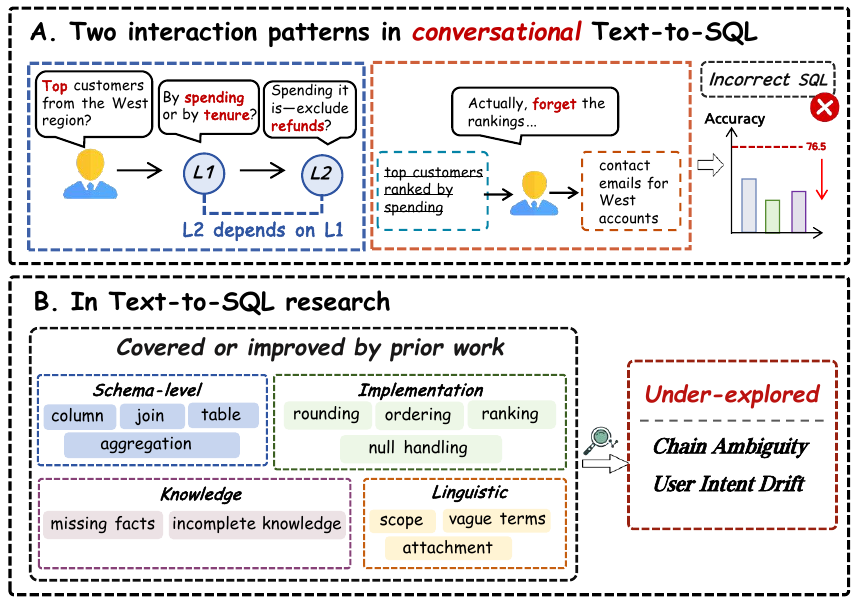}
    \vspace{-6mm}
    \caption{(A) Two conversational patterns that challenge text-to-SQL. (B) Neither is covered by existing text-to-SQL benchmarks or standard evaluation protocols.}
    \label{fig:introduction}
    \vspace{-5mm}
\end{figure}

However, \textbf{how does user intent actually unfold and shift across turns, and are current agents equipped to handle it?}
Figure~\ref{fig:introduction} illustrates two recurring interaction patterns in multi-turn dialogue. First, an underspecified question may trigger \textbf{chain ambiguity}: a layered clarification process in which a later clarification is conditionally dependent on the resolution of an earlier one, going beyond turn-local ambiguity studied in prior work~\cite{bhaskar2023ambiqt,saparina2024ambrosia,dong2025practiq}. Second, a user may retract and replace a committed request element mid-conversation, a non-monotonic shift we term \textbf{user intent drift}. These two interaction patterns raise two critical concerns:
\textbf{1) Chain dependency is not captured by turn-local ambiguity}, as resolving each clarification in isolation says nothing about whether the agent recognizes the dependency between layers;
\textbf{2) Commitment dynamics are not captured by end-to-end accuracy}, as a single execution match cannot tell whether the agent registered the retraction or merely happened to land on the right query.

Existing conversational text-to-SQL benchmarks~\cite{yu2019sparc,yu2019cosql,sun2025dysql,huo2026interact} measure multi-turn SQL accuracy, while a related line of work~\cite{bhaskar2023ambiqt,saparina2024ambrosia,dong2025practiq,ding2026ambisql} also defines turn-local ambiguity taxonomies for individual utterances.
However, these efforts either evaluate dialogue execution as a single end-to-end number or restrict ambiguity to one turn at a time, leaving open how agents handle cross-turn dependency and retraction. In this context, \textbf{effectively evaluating how text-to-SQL agents handle these dialogue-level interaction patterns remains a significant challenge}. As a solution, this paper introduces \textbf{\textsc{TIDE-Bench}}, a benchmark for conversa\textbf{\textsc{ti}}onal text-to-SQL under chain ambiguity and intent \textbf{\textsc{d}}rift \textbf{\textsc{e}}valuation, which operationalizes chain ambiguity through a counterfactual test for conditional dependency and intent drift through a non-monotonicity test over commitment sets derived from SQL abstract syntax trees.

Specifically, \textsc{TIDE-Bench} is built around a paired factorial design over 514 anchor SQLs from BIRD~\cite{li2023bird}: each anchor spawns three samples activating chain ambiguity alone and intent drift alone and together, so that performance differences are attributable to the activated interaction patterns rather than others. The resulting 1,542 samples are produced through a four-stage pipeline with automated validation and human verification. Beyond execution accuracy, \textsc{TIDE-Bench} introduces dedicated metrics for chain-layer identification and drift recognition-resolution. Overall, our main contributions are summarized as follows:
\vspace{-1mm}
\begin{itemize}
\item We identify and formalize \textbf{chain ambiguity} and \textbf{user intent drift} as two underexamined dialogue-level interaction patterns in conversational text-to-SQL, and operationalize each through a counterfactual test for conditional dependency and a non-monotonicity test over commitment sets for systematic evaluation.
\vspace{-2mm}
\item We introduce \textbf{\textsc{TIDE-Bench}}, a benchmark of 514 anchor SQLs and 1,542 paired samples constructed through a four-stage pipeline with automated validation and human verification, whose design attributes performance differences to the activated interaction patterns.
\vspace{-2mm}
\item We evaluate 12 advanced LLMs on \textsc{TIDE-Bench} and also diagnose three key distinct failure modes: a persistent chain identification bottleneck unaffected by ask behavior, a drift recognition-resolution gap concentrated on retraction-target inference, and a joint penalty where the two failures occur on overlapping rather than independent samples.
\end{itemize}
\section{Problem Formalization}
\label{sec:formalization}

Real-world text-to-SQL systems often operate in multi-turn interactions, where a user's intent may be underspecified at first and clarified over time. During this process, the intended SQL query may evolve as the user resolves ambiguity or revises a prior choice. We formalize two key interaction patterns:
\textbf{chain ambiguity} and \textbf{user intent drift}.

\subsection{Chain Ambiguity}
\label{subsec:chain} 

\noindent\textbf{Definition.} A sample exhibits \emph{chain ambiguity} when an underspecified user query induces two layers of clarification: \textbf{L1 (layer one)} and \textbf{L2 (layer two)}, such that the existence, scope, or framing of L2 is conditionally dependent on the resolution of L1. Each layer offers two semantically plausible interpretations of the query. This conditional dependency distinguishes \emph{chain ambiguity} from two stacked independent ambiguities, where simply resolving L1 leaves L2's semantic content unaffected.

\namedpara{Counterfactual Test.} To distinguish chain ambiguity from independent stacking, we use the following
\emph{counterfactual test}~\citep{pearl2009causality}: if L1 were resolved with its alternative choice, would L2 still be posed as the same question? The test yields one of three verdicts: (i) \textbf{Eliminated}, where L2's premise vanishes under the L1 alternative and the question becomes vacuous; (ii) \textbf{Transformed}, where L2 must be reframed through changes to its referent, vocabulary, or scope; and (iii) \textbf{Invariant}, where L2 can be asked verbatim, indicating semantic independence. A sample qualifies as chain-ambiguous only when the verdict is \emph{eliminated} or \emph{transformed}; samples judged \emph{invariant} are rejected at this stage.

\begin{figure*}[!t]
  \centering
  \includegraphics[width=\textwidth]{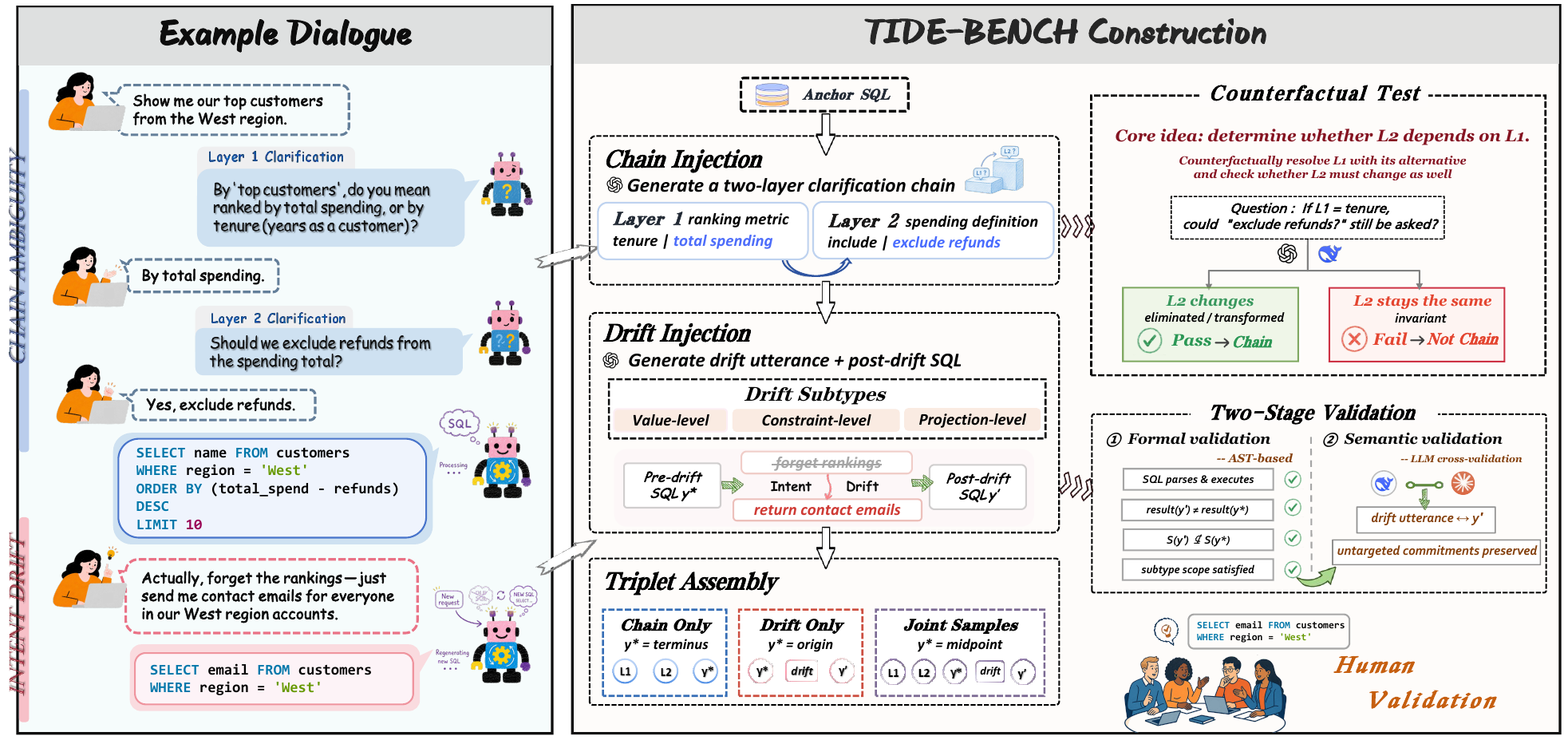}
  \vspace{-6mm}
  \caption{\textbf{Left:} motivating dialogue illustrating chain ambiguity (L1/L2 clarification with conditional dependency) and user intent drift (non-monotonic retraction of a prior commitment). \textbf{Right:} overview of the \textsc{TIDE-Bench} four-stage construction pipeline (anchor selection, chain injection, drift injection, and triplet assembly; see \S\ref{sec:construction}).}
  \label{fig:construction}
  \vspace{-4mm}
\end{figure*}

\namedpara{Worked Example.}
For the dialogue in Figure~\ref{fig:construction}, the L1 alternative
\emph{tenure} measures duration rather than a monetary quantity, so L2's question about excluding refunds no longer has a valid referent. The verdict for this example is therefore \emph{eliminated}.

\subsection{User Intent Drift}
\label{subsec:drift}

\noindent\textbf{Definition.}
\emph{User intent drift} is a non-monotonic revision in which at least one previously established commitment is withdrawn or replaced. Unlike cooperative continuation, which retains all prior commitments, drift requires revising the dialogue state and discarding withdrawn requirements from the resulting query but may also introduce new commitments. To formalize this definition, the user's committed intent at each dialogue turn is represented by a SQL query $y$ encoding the intent accumulated up to that turn and a corresponding \emph{commitment set} $S(y)$ directly extracted from its full abstract syntax tree. The set covers filtering predicates, projections, grouping and ordering keys, aggregation filters, and row-limit and offset specifications (Appendix~\ref{app:commitment-extractor}). Let
$S_{\mathrm{pre}}=S(y_{\mathrm{pre}})$ and $S_{\mathrm{post}}=S(y_{\mathrm{post}})$.

\namedpara{Non-Monotonicity Test.}
Drift occurs iff $S_{\mathrm{pre}}\nsubseteq S_{\mathrm{post}}$. The test itself places no restriction on added commitments,
$S_{\mathrm{post}}\setminus S_{\mathrm{pre}}$, and allows removal and replacement while excluding no-change updates and pure additions. Each dialogue contains one controlled drift event, which may affect multiple commitments. This design keeps the pre- and post-drift targets identifiable under the paired design.

\namedpara{Drift Subtypes.}
The benchmark distinguishes three subtypes: (i) \textbf{Value-level}, which replaces only one predicate literal while preserving predicate structure; (ii) \textbf{Constraint-level}, which removes one or more filtering predicates and makes no other uery changes; and (iii) \textbf{Projection-level}, which changes only output-side commitments while preserving all filtering predicates. Algorithm~\ref{alg:drift-test} operationalizes this decision process, first applying the non-monotonicity test and then assigning each accepted drift update to one of the three subtypes.
\section{\textsc{TIDE-Bench} Construction}
\label{sec:construction}

We describe the construction of \textsc{TIDE-Bench}. The benchmark is built around a paired design that isolates each interaction pattern from the anchor.

\subsection{Paired Factorial Design}
\label{subsec:paired-factorial}

\textsc{TIDE-Bench} uses a \textbf{paired factorial design}: each BIRD anchor SQL $y^*$ that passes construction filters spawns three samples sharing the anchor but activating different interaction patterns: $\mathcal{C}_{y^*}$ (Chain-only), $\mathcal{D}_{y^*}$ (Drift-only), and $\mathcal{J}_{y^*}$ (Joint). Pairing by anchor controls for schema complexity, domain difficulty, and surface ambiguity, attributing performance differences to the activated patterns rather than any uncontrolled anchor-specific confounds.

\subsection{Construction Pipeline}
\label{subsec:pipeline}

Building on this design, our construction pipeline proceeds in four stages: anchor selection, chain injection, drift injection, and triplet assembly (Figure~\ref{fig:construction}, right). We summarize each stage below; full prompts, judge configurations, and associated validation checks are instead deferred to Appendix~\ref{app:pipeline}.

\begin{algorithm}[t]

\caption{User Intent Drift Operationalization}
\label{alg:drift-test}
\small
\renewcommand{\arraystretch}{1.10}

\begin{tabularx}{\linewidth}{
    @{}r@{\hspace{0.75em}}
    >{\raggedright\arraybackslash}X@{}
}

\multicolumn{2}{@{}l}{
    \textbf{Require:} Queries
    $y_{\mathrm{pre}}$ and $y_{\mathrm{post}}$
} \\

\multicolumn{2}{@{}l}{
    \textbf{Ensure:} Drift status or subtype
} \\[2pt]

\multicolumn{2}{@{}l}{
    \textbf{Notation:}
} \\[0pt]

&
$\mathcal{W}(y)$: filtering commitments in $y$
\\

&
$\mathcal{O}(y)=S(y)\setminus\mathcal{W}(y)$:
output-side commitments in $y$
\\

&
$\operatorname{LitReplace}
(\mathcal{W}_{\mathrm{pre}},\mathcal{W}_{\mathrm{post}})$:
one predicate literal changes; its structure and all other filters
remain fixed
\\[2pt]

&
\textcolor{gray}{\itshape\% Extract commitments}
\\[0pt]

1: &
$S_{\mathrm{pre}}\gets S(y_{\mathrm{pre}})$;
$S_{\mathrm{post}}\gets S(y_{\mathrm{post}})$
\\

2: &
Set $\mathcal{W}_t\gets\mathcal{W}(y_t)$ and
$\mathcal{O}_t\gets\mathcal{O}(y_t)$ for
$t\in\{\mathrm{pre},\mathrm{post}\}$
\\[2pt]

&
\textcolor{gray}{\itshape\% Test non-monotonicity}
\\[0pt]

3: &
\textbf{if} $S_{\mathrm{pre}}\subseteq S_{\mathrm{post}}$
\textbf{then}
\\

4: &
\hspace*{1.2em}\textbf{return} \textsc{Not Drift}
\\

5: &
\textbf{end if}
\\[2pt]

&
\textcolor{gray}{\itshape\% Assign drift subtype}
\\[0pt]

6: &
\textbf{if}
$\mathcal{O}_{\mathrm{pre}}=\mathcal{O}_{\mathrm{post}}$
and
$\operatorname{LitReplace}
(\mathcal{W}_{\mathrm{pre}},\mathcal{W}_{\mathrm{post}})$
\textbf{then}
\\

7: &
\hspace*{1.2em}\textbf{return} \textsc{Value-level}
\\

8: &
\textbf{end if}
\\

9: &
\textbf{if}
$\mathcal{O}_{\mathrm{pre}}=\mathcal{O}_{\mathrm{post}}$
and
$\mathcal{W}_{\mathrm{post}}
\subsetneq\mathcal{W}_{\mathrm{pre}}$
\textbf{then}
\\

10: &
\hspace*{1.2em}\textbf{return} \textsc{Constraint-level}
\\

11: &
\textbf{end if}
\\

12: &
\textbf{if}
$\mathcal{W}_{\mathrm{pre}}=\mathcal{W}_{\mathrm{post}}$
and
$\mathcal{O}_{\mathrm{pre}}
\nsubseteq\mathcal{O}_{\mathrm{post}}$
\textbf{then}
\\

13: &
\hspace*{1.2em}\textbf{return} \textsc{Projection-level}
\\

14: &
\textbf{end if}
\\

15: &
\textbf{return} \textsc{Reject}
\\

\end{tabularx}

\end{algorithm}

\namedpara{Anchor selection.} Candidate anchors are drawn from BIRD~\citep{li2023bird}, retaining only SQLs that (i) execute on their SQLite database, (ii) carry enough structural complexity for two-layer chain injection, and (iii) admit at least one drift subtype from \S~\ref{subsec:drift}. Stratified sampling balances complexity tiers and source splits, totaling \textbf{830 anchors}.

\namedpara{Chain injection.} For each anchor, an advanced LLM (GPT-5.1) generates the two-layer clarification structure $(L1, L2)$, an underspecified initial question, chosen responses, and unchosen alternatives, conditioned on the anchor SQL as the canonical $L1 + L2$ target. Two independent judges (GPT-5.3-Codex, DeepSeek-V4-Pro) apply the counterfactual test (\S~\ref{subsec:chain}) and discard \emph{invariant} candidates. This stage retains 695/830 candidates (83.7\%), with 94.7\% overall inter-judge agreement. A final LLM rewrite of BIRD's evidence field removes clauses pre-resolving L1 or L2 while faithfully preserving schema-essential mappings (Appendix~\ref{app:pipeline}).

\namedpara{Drift injection.} For each chain-validated anchor, the same generator produces a drift utterance and post-drift target $y_{\mathrm{post}}^*$ conditioned on a pre-assigned subtype. Each pair passes two stages. AST-level validation enforces parseability, executability, result distinctness, the non-monotonicity condition $S_{\mathrm{pre}} \nsubseteq S_{\mathrm{post}}$, and subtype-specific scope constraints. LLM cross-validation (Claude-Sonnet-4.6, DeepSeek-V4-Pro) assesses semantic validity beyond formal checks, namely faithfulness to the drift turn and preservation of non-targeted structural elements. The two stages are complementary: AST checks enforce the definition, while cross-validation guards against drifts pass checks but misalign with semantic intent. AST validation retains 554/695 (79.7\%); cross-validation retains 514/554 (92.8\%), with 96.6\% overall inter-judge agreement.

\namedpara{Triplet assembly.} Surviving chain and drift constructions are paired by anchor to form \textbf{514 triplets (1{,}542 samples)}, each consisting of matched Chain-only, Drift-only, and Joint instances built over the same anchor SQL and sharing the same schema. This pairing supports controlled performance comparisons across all the three interaction settings.

\subsection{Human Validation}
\label{sec:human-validation}

Beyond the automated filters above, we additionally verify the quality and construct validity of \textsc{TIDE-Bench} through human annotation. We conduct a stratified audit over 250 triplets (750 samples), balanced across sample type, drift subtype, counterfactual verdict, and SQL complexity tier. Each sample is independently annotated by five annotators with NLP and production SQL backgrounds, blind to automated verdicts produced during construction. Rather than a holistic judgment, we use a per-item checklist that operationalizes the definitions in \S~\ref{sec:formalization}; the full checklist and annotator details appear in Appendix~\ref{app:validation} for reproducibility.

We report \emph{majority acceptance} ($\geq$3/5 annotators pass every applicable item) and \emph{strict acceptance} (unanimity). Overall majority and strict acceptance are 92.4\% and 83.1\% (Table~\ref{tab:human-validation}; all Fleiss' $\kappa > 0.77$). Counterfactual dependency and non-monotonic retraction pass on 93.4\% and 94.8\% respectively, while joint transition coherence is slightly lower at 91.6\%, as it requires both interaction patterns to be validated jointly in the dialogue.

\begin{table}[!t]
\centering
\small
\setlength{\tabcolsep}{3pt}
\renewcommand{\arraystretch}{1.35}

\begin{tabular*}{\columnwidth}{
    @{\extracolsep{\fill}}cccc@{}
}
\toprule
\textbf{Split}
& \textbf{Majority Acc.}
& \textbf{Strict Acc.}
& \textbf{Fleiss' $\boldsymbol{\kappa}$} \\
\midrule
\textsc{Chain-only} & 93.2 & 84.8 & 0.82 \\
\textsc{Drift-only} & 95.6 & 87.6 & 0.80 \\
\textsc{Joint}      & 88.4 & 76.8 & 0.77 \\
\midrule
Overall             
& \textbf{92.4}
& \textbf{83.1}
& \textbf{0.81} \\
\bottomrule
\end{tabular*}

\vspace{-1mm}
\caption{Human validation on 250 stratified triplets. Majority requires $\geq 3/5$ annotators per item; strict requires unanimity. Fleiss' $\kappa$ uses binary accept/reject verdicts.}
\label{tab:human-validation}
\vspace{-5mm}
\end{table}
\begin{table*}[t]
\centering
\normalsize
\setlength{\tabcolsep}{3pt}
\renewcommand{\arraystretch}{1.35}

\begin{tabularx}{\linewidth}{@{}p{0.08\linewidth} p{0.32\linewidth} p{0.47\linewidth} p{0.07\linewidth}@{}}
\toprule
\textbf{Metric}
& \textbf{Full Name}
& \textbf{Definition}
& \textbf{Set} \\
\midrule
DEA & Dialogue Execution Accuracy
& $\frac{1}{|\mathcal{S}|} \sum_{\tau \in \mathcal{S}} \prod_{i=1}^{n_\tau} \mathrm{EX}(y_i, y_i^*)$
& all \\

AR & Ask Rate
& $\frac{1}{|\mathcal{S}|} \sum_{\tau \in \mathcal{S}} \mathds{1}\!\left[\tau \text{ contains a clarification}\right]$
& all \\

L1-IR & L1 Identification Rate
& $\frac{1}{|\mathcal{S}|} \sum_{\tau \in \mathcal{S}} \mathds{1}\!\left[\exists\, t : \rho_t = \mathrm{L1}\right]$
& $\mathcal{C}, \mathcal{J}$ \\

L2-IR$^{p}$ & L2 Identification Rate
& $\frac{1}{|\mathcal{S}|} \sum_{\tau \in \mathcal{S}} \mathds{1}\!\left[\exists\, t : \rho_t = \mathrm{L2}\right]$
& $\mathcal{C}, \mathcal{J}$ \\

JIR & Joint Identification Rate
& $\frac{1}{|\mathcal{S}|} \sum_{\tau \in \mathcal{S}} \mathds{1}\!\left[\exists\, t_1, t_2 : \rho_{t_1} = \mathrm{L1} \wedge \rho_{t_2} = \mathrm{L2}\right]$
& $\mathcal{C}, \mathcal{J}$ \\

DRR & Drift Recognition Rate
& $\frac{1}{|\mathcal{S}|} \sum_{\tau \in \mathcal{S}} \mathds{1}\!\left[\mathrm{Rec}(\tau)\right]$
& $\mathcal{D}, \mathcal{J}$ \\

DRA & Drift Resolution Accuracy
& $\frac{1}{|\mathcal{S}|} \sum_{\tau \in \mathcal{S}} \mathds{1}\!\left[\mathrm{Rec}(\tau) \wedge \mathrm{EX}(y_{\mathrm{post}}, y_{\mathrm{post}}^*)\right]$
& $\mathcal{D}, \mathcal{J}$ \\
\bottomrule
\end{tabularx}
\vspace{-1mm}
\caption{\textsc{TIDE-Bench} metrics spanning four behavioral facets: execution accuracy (DEA), clarification behavior (AR), chain-layer identification (L1-IR, L2-IR$^{p}$, JIR), and drift handling (DRR, DRA). $\mathcal{C}, \mathcal{D}, \mathcal{J}$ denote Chain-only, Drift-only, and Joint sample sets. L2-IR$^{p}$ uses the $p$-protocol (\S~\ref{subsec:interaction}); all other metrics use the $f$-protocol instead.}
\label{tab:metrics}
\vspace{-4mm}
\end{table*}
\subsection{Dataset Statistics}
\label{sec:dataset-statistics}
\textsc{TIDE-Bench} comprises 514 triplets and 1{,}542 benchmark samples (514 each of Chain-only, Drift-only, Joint). Anchors are partitioned into basic, intermediate, and advanced tiers by a rule-based score combining anchor-SQL structure with chain- and drift-induced factors. Drift subtypes are evenly covered: value level 34.0\%, constraint-level 31.3\%, and projection-level 34.6\%. Counterfactual verdicts are balanced between \emph{eliminated} (49.4\%) and \emph{transformed} (50.6\%), while the overall complexity is mostly intermediate (61.3\%), with basic and advanced tiers at 26.5\% and 12.3\%, respectively.

\section{Evaluation Protocol}

\label{sec:protocol}
Building on the construction in \S~3, this section specifies how each \textsc{TIDE-Bench} sample is turned into an evaluation run for the experiments below.

\subsection{Interaction Setup}
\label{subsec:interaction}

Each \textsc{TIDE-Bench} sample defines a dialogue task where an agent interacts with a user simulator (\S~\ref{subsec:simulator}) over a database $\mathcal{D}$ with schema $\mathcal{S}$. The dialogue begins with the user's initial question $u_1$ and proceeds in alternating turns $(u_1, a_1, u_2, a_2, \ldots)$. Each agent turn $a_i$ is either a clarification request or candidate SQL query against $\mathcal{D}$; the simulator returns $u_{i+1}$. For Drift-only and Joint samples, the harness injects the pre-annotated drift turn verbatim after the agent's first SQL query, ensuring deterministic drift content across evaluation runs.

\namedpara{\(\boldsymbol{f}\)-protocol and \(\boldsymbol{p}\)-protocol.} Each sample is evaluated under two protocols. The \textbf{\(\boldsymbol{f}\)-protocol} (\emph{full}) starts from $u_1$, requiring the agent to elicit and resolve both clarification layers for Chain-only and Joint samples. The \textbf{\(\boldsymbol{p}\)-protocol} (L1-\emph{primed}) pre-populates the history with a synthetic three-turn L1 resolution: the initial question, annotated L1 clarification, and canonical L1 answer. It marks the simulator's L1 state as resolved before the first live turn. The $p$-protocol isolates L2, testing whether agents miss it once L1 is no longer a confound. Full harness parameter settings appear in Appendix~\ref{app:evaluation-setup}.

\subsection{User Simulator}
\label{subsec:simulator}

The simulator follows the two-stage function-driven design of recent advance~\cite{huo2026interact}, with three minor adaptations for \textsc{TIDE-Bench}'s two-layer chain structure (Appendix~\ref{app:user-sim}). The router decisions $\rho_t$ used by  the metrics below are produced at each user turn responding to an agent clarification. We validate the simulator on USERSIM-GUARD and on agent--simulator traces independently labeled by five SQL experts (Fleiss' $\kappa = 0.802$); full configuration provided in Appendix~\ref{app:user-sim}.

\subsection{Metrics}
\label{subsec:metrics}

Seven metrics span four behavioral facets: execution accuracy, clarification behavior, chain-layer identification and drift handling. All are computed as proportions over a sample set $\mathcal{S}$ of a given type.

\namedpara{Notation.}
Let $\tau$ denote a dialogue trajectory containing $n_\tau \in \{1, 2\}$ SQL queries $y_1, \ldots, y_{n_\tau}$ paired with $y_1^*, \ldots, y_{n_\tau}^*$. Chain-only samples have $n_\tau = 1$, while Drift-only and Joint samples have $n_\tau = 2$, with $y_1 = y_{\mathrm{pre}}$ and $y_2 = y_{\mathrm{post}}$. Let
$\mathrm{EX}(y, y^*) \in \{0, 1\}$ denote execution match, i.e., equivalent result sets modulo row order; $\rho_t \in \{\mathrm{L1}, \mathrm{L2}, \mathrm{Other}\}$ denote the router decision (\S~\ref{subsec:simulator}) for the agent's clarification at turn $t$; and $S(y)$ denote the commitment set of $y$ (\S~\ref{subsec:drift}). Define $\mathrm{Rec}(\tau) := \bigl[ S(y_{\mathrm{pre}}) \nsubseteq S(y_{\mathrm{post}}) \bigr]$ as the non-monotonicity condition used for drift recognition.

\begin{table*}[t]
\centering
\small
\setlength{\tabcolsep}{3pt}
\renewcommand{\arraystretch}{1.52}

\begin{tabular*}{\linewidth}{
@{}
c@{\hspace{4pt}}
l@{\hspace{6pt}}
c@{\extracolsep{\fill}}
c c c c c c c c
@{}
}

\toprule
\multirow{2}{*}{\textbf{\#}}
& \multicolumn{1}{c@{\hspace{6pt}}}
  {\multirow{2}{*}{\textbf{Model}}}
& \multirow{2}{*}{\textbf{Chain-only}}
& \multicolumn{4}{c}{\textbf{Drift-only}}
& \multicolumn{4}{c}{\textbf{Joint}} \\
\cmidrule(lr){4-7}
\cmidrule(lr){8-11}

& &
& \textbf{Value}
& \textbf{Constraint}
& \textbf{Projection}
& \textbf{All}
& \textbf{Value}
& \textbf{Constraint}
& \textbf{Projection}
& \textbf{All} \\

\midrule
1  & Claude-Sonnet-4.6 & \textbf{35.9} & \textbf{47.7} & \textbf{44.2} & 43.4 & \textbf{45.1} & \textbf{31.2} & 20.2 & \textbf{31.4} & \textbf{27.8} \\
2  & Qwen2.5-72B       & 32.4 & 44.3 & 41.1 & \underline{44.0} & \underline{43.2} & \underline{30.1} & \underline{21.5} & 28.6 & \underline{26.8} \\
3  & o4-mini           & \underline{32.9} & 42.6 & 41.1 & 40.0 & 41.2 & \textbf{31.2} & \underline{21.5} & 27.4 & \underline{26.8} \\
4  & GPT-4o            & 30.9 & \underline{46.0} & 38.7 & \textbf{44.6} & \underline{43.2} & 27.8 & \textbf{22.1} & \underline{29.7} & 26.7 \\
5  & GPT-5.4           & 32.5 & 43.2 & 36.8 & 42.3 & 40.9 & 27.8 & 20.9 & \underline{29.7} & 26.3 \\
6  & Llama-3.3-70B     & 32.7 & 40.3 & 39.3 & 38.9 & 39.5 & 29.5 & \textbf{22.1} & 26.9 & 26.3 \\
7  & o3-mini           & 32.6 & 40.9 & 37.4 & 38.9 & 39.1 & 29.5 & 18.4 & 28.6 & 25.7 \\
8  & Gemini-2.5-Flash  & 31.6 & 39.2 & 42.9 & 33.1 & 38.3 & 27.3 & 20.2 & 27.4 & 25.1 \\
9  & Grok-4.1-Fast     & 30.4 & 41.5 & 36.2 & 41.7 & 39.9 & 29.0 & 18.4 & 25.1 & 24.3 \\
10 & Qwen-Max          & 30.5 & 40.3 & \underline{43.6} & 40.6 & 41.4 & 28.4 & 20.9 & 21.7 & 23.7 \\
11 & DeepSeek-V4-Flash & 27.3 & 40.9 & 39.9 & 39.4 & 40.1 & 21.6 & 17.8 & 26.9 & 22.2 \\
12 & Grok-4-Fast       & 25.9 & 36.9 & 33.7 & 38.9 & 36.6 & 19.9 & 14.7 & 22.9 & 19.3 \\
\bottomrule

\end{tabular*}

\vspace{-1mm}
\caption{DEA (\%) under the paired factorial design. The best and second-best results per column are in bold and underlined, respectively. Chain-only reports strict execution accuracy; Drift-only and Joint report end-to-end accuracy broken down by drift subtype; All denotes aggregate performance, and higher metric values are better.}
\label{tab:dea}
\vspace{-1mm}
\end{table*}
\begin{figure*}[!t]
  \centering
  \includegraphics[width=\textwidth]{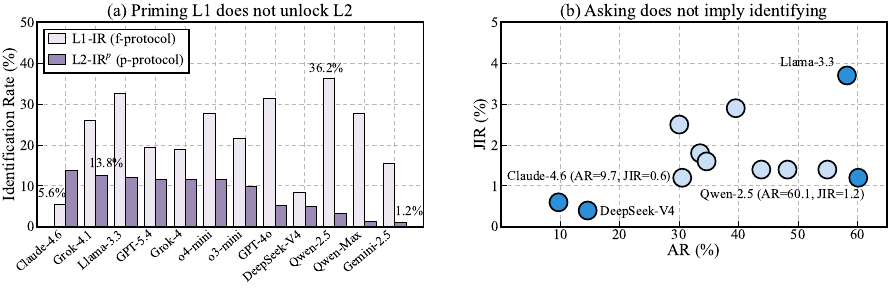}
  \vspace{-7mm}
  \caption{Chain-layer identification on Chain-only samples. (a) L1-IR ($f$-protocol) vs.\ L2-IR$^p$ ($p$-protocol) per model: even with L1 pre-resolved, L2 identification stays low. (b) JIR against ask rate: identification remains low regardless of how much the agent asks. Model names are abbreviated; full names appear in Table~\ref{tab:dea} for reference.}
  \label{fig:chain}
  \vspace{-5mm}
\end{figure*}

\namedpara{Metric definitions.} The seven metrics are defined in Table~\ref{tab:metrics}. \textbf{DEA} requires every SQL query in a trajectory to pass execution match. \textbf{AR} measures whether the agent asks, regardless of layer, providing a baseline clarification rate. \textbf{L1-IR}, \textbf{L2-IR$^{p}$}, and \textbf{JIR} measure whether the agent surfaces clarification requests routed to the corresponding layer within a trajectory, with L2-IR$^{p}$ evaluated under the $p$-protocol to isolate L2 from L1 confounds. \textbf{DRR} captures whether the agent's commitment sets exhibit the required non-monotonic change after the drift utterance, and \textbf{DRA} additionally requires the resulting post-drift SQL to be executionally correct.

\section{Experimental Results}

\label{sec:experiment}

We evaluate 12 advanced LLMs on \textsc{TIDE-Bench}. We report model accuracy across the three sample types, diagnose where chain ambiguity and intent drift individually fail, and examine their interaction when both patterns co-occur in a single dialogue. This analysis distinguishes recognition from resolution and reveals how errors propagate across dialogue phases into their final SQL predictions.

\subsection{Overall Performance}
\label{subsec:performance}

Table~\ref{tab:dea} reports DEA across the three sample types. Performance is low throughout: even the strongest evaluated model, Claude-Sonnet-4.6, reaches only 45.1\% on Drift-only and drops to 27.8\% on Joint, and no model exceeds 46\% overall on any sample type. DEA is consistently highest on Drift-only, lowest on Joint, and intermediate on Chain-only.

This consistently low performance is not concentrated in any particular model class. Larger evaluated models outperform smaller by only 2--3 percentage points on average across three conditions, and the spread between top and bottom models stays within roughly 10 percentage points everywhere. The overall low performance therefore appears to come less from overall model scale than from how agents handle chain ambiguity and user intent drift. The following analyses separate these errors into chain identification, drift recognition, and drift resolution across models and subtypes, and examine how they compound in Joint samples.

\begin{figure*}[!t]
  \centering
  \includegraphics[width=\textwidth]{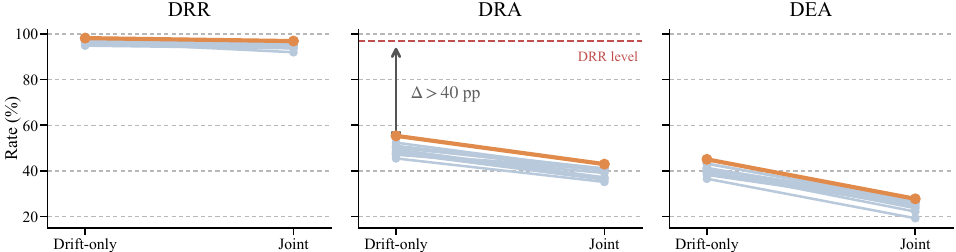}
  \vspace{-6mm}
  \caption{Drift handling from Drift-only to Joint across DRR, DRA, and DEA, with one line per model. DRR remains high and flat, while DRA and DEA start lower and decline further in Joint. The dashed DRA reference marks DRR, highlighting the recognition--resolution gap that persists for all models across both interaction settings. The orange line highlights Claude-Sonnet-4.6, which achieves the highest DEA in both Drift-only and Joint.}
  \label{fig:slope}
  \vspace{-2mm}
\end{figure*}

\subsection{What Makes Chain Challenging}
\label{subsec:chain-bottleneck}

\noindent\textbf{Layer-level identification.}
Resolving a chain-only sample requires recognizing dependent clarification layers from the initial question. We measure this directly through per-layer identification rates, with per-model details in Table~\ref{tab:chain-identification} (Appendix~\ref{app:chain-id}). JIR stays below 5\% for every model and below 2\% for the majority, \textbf{\textit{showing that no evaluated model reliably surfaces both clarification layers}}. L1-IR ranges from 5.6\% (Claude-Sonnet-4.6) to 36.2\% (Qwen2.5-72B), with most models clustered in the 20--30\% range, despite L1 being directly available in the initial question. One plausible explanation for the even lower JIR is that L1 difficulty cascades downward: when L1 is missed, the agent may never reach L2 at all. The $p$-protocol (\S~\ref{subsec:interaction}) is designed to remove exactly this confound, pre-populating L1 with its canonical resolution so that L2 is isolated as the sole downstream identification target. Figure~\ref{fig:chain}(a) contrasts L1-IR under the $f$-protocol with L2-IR$^p$ under the $p$-protocol for each model. Even with L1 externally resolved, L2-IR$^p$ ranges only from 1.2\% (Gemini-2.5-Flash) to 13.8\% (Claude-Sonnet-4.6), with half the models still below 10\%, \textbf{\textit{showing that L2 identification fails on its own terms rather than as a side effect of L1 difficulty}}.

\begin{figure}[t]
  \centering
  \includegraphics[width=\columnwidth]{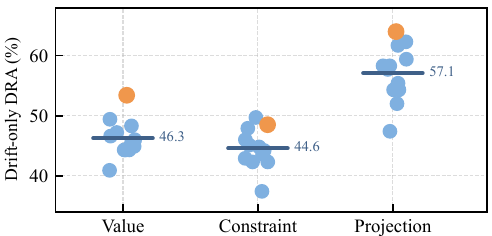}
  \vspace{-5mm}
  \caption{Drift-only DRA by drift subtype. Points denote models, and the bar marks the corresponding subtype mean. Orange points mark per-subtype maxima.}
  \label{fig:subtype}
  \vspace{-5mm}
\end{figure}

\namedpara{Ask behavior.}
Could this low identification simply reflect how agents ask? An agent that rarely asks for clarification at all, or one that asks freely without targeting genuine ambiguity, would both depress identification rates. Figure~\ref{fig:chain}(b) plots JIR against ask rate (AR) for all 12 models. Ask rates vary widely, ranging from below 10\% to over 60\%, yet JIR stays under 4\% across the entire range. \textbf{\textit{Models that ask more do not surface the chain more reliably}}. Claude-Sonnet-4.6 and Qwen2.5-72B illustrate the two extremes: Claude asks on only 9.7\% of Chain-only samples but Qwen2.5-72B asks on 60.1\%, yet their JIR are at most 1.2\% (0.6\% and 1.2\% respectively). To further isolate whether high asking is selective or indiscriminate, we measure the ask rate on Drift-only samples, where the initial user question carries no chain ambiguity, so any clarification reflects asking that is not ambiguity-driven. Claude's Drift-only ask rate (Table~\ref{tab:chain-identification}) drops to 0.4\%, while Qwen2.5-72B's remains 20.0\%, confirming opposite ask styles. \textbf{\textit{Whether an agent asks selectively or indiscriminately, the chain remains unidentified, pointing to identification capacity itself as the bottleneck}}.

\subsection{Where Drift Handling Fails}
\label{subsec:drift-bottleneck}

\noindent\textbf{Recognition--resolution gap.}
While chain failure is primarily an identification problem, drift exhibits a different failure pattern. We measure whether agents register the drift turn (DRR) and whether they recover an executionally correct post-drift SQL afterward (DRA). Figure~\ref{fig:slope} presents both metrics, with per-model values in Appendix~\ref{app:chain-id}. DRR is at least 94.9\% across models, while DRA remains roughly within the 45--55\% band, \textbf{\textit{leaving every evaluated model with a gap exceeding 40 percentage points between recognizing the drift and successfully resolving user's revised request}}.

\begin{table}[!t]
\centering
\footnotesize
\linespread{1.05}\selectfont
\renewcommand{\arraystretch}{1.2}
\begin{tabularx}{\linewidth}{@{}>{\raggedright\arraybackslash}X@{}}
\toprule

\casehead{Context}\par
\vspace{2pt}
\textit{Schema:}
\texttt{coaches(coachID, year, tmID, won, lost)};
\texttt{teams(year, tmID, won, lost)}.\par
\textit{Question:}
``Which coaches had disappointing seasons in 2000--2010?
Show their coach ID, team, and year.''
\\

\midrule

\casehead{Chain Ambiguity}\par
\vspace{2pt}
\textit{Gold path:}
$L1$: overall losing record
$\rightarrow$
$L2$: team's full-season record.\par
{\color{black!60}
\textit{Alternatives:}\ 
\mbox{$L1$: away-game record;}\ 
\mbox{$L2$: coach-stint record.}}
\\

\addlinespace[2.5pt]
\sqlbox{Gold SQL:}{SELECT DISTINCT c.coachID, t.tmID, c.year FROM coaches c JOIN teams t ON c.tmID=t.tmID AND c.year=t.year WHERE c.year BETWEEN 2000 AND 2010 AND t.lost>t.won}
\\

\addlinespace[3pt]
\textbf{Claude-Sonnet-4.6:} Clarifies that ``disappointing'' means a losing record, but uses
\sql{coaches.lost>coaches.won} without asking whose record should
define the season.\par
\vspace{1pt}
{\scriptsize
\color{black!60}
$L1$ recognized
\quad
$L2$ missed
\quad
\textcolor{red!75!black}{preEX=False \ding{55}}}
\\

\addlinespace[2.5pt]
\textbf{Llama-3.3-70B:}
Also resolves the losing-record interpretation, but applies
\sql{coaches.lost > coaches.won} despite joining \texttt{teams}.\par
\vspace{1pt}
{\scriptsize
\color{black!60}
$L1$ recognized
\quad
$L2$ missed
\quad
\textcolor{red!75!black}{preEX=False \ding{55}}}
\\

\midrule

\casehead{User Intent Drift}\,
{\scriptsize\color{black!60}(constraint-remove)}\par
\vspace{2pt}
Drift utterance: ``Do not limit it to 2000--2010; show those coaches for all years.''
\\

\addlinespace[2.5pt]
\sqlbox{Gold post-drift SQL:}{SELECT DISTINCT c.coachID, t.tmID, c.year FROM coaches c JOIN teams t ON c.tmID=t.tmID AND c.year=t.year WHERE t.lost>t.won}
\\

\addlinespace[3pt]
\textbf{Claude-Sonnet-4.6:}
Removes the year predicate but retains the coach-stint criterion.\par
\vspace{1pt}
{\scriptsize
\color{black!60}
\textcolor{green!50!black}{DRR=True \ding{51}}
\quad
chain error inherited
\quad
\textcolor{red!75!black}{postEX=False \ding{55}}}
\\

\addlinespace[2.5pt]
\textbf{Llama-3.3-70B:}
Makes the same drift update while retaining the same chain error.\par
\vspace{1pt}
{\scriptsize
\color{black!60}
\textcolor{green!50!black}{DRR=True \ding{51}}
\quad
chain error inherited
\quad
\textcolor{red!75!black}{postEX=False \ding{55}}}
\\

\bottomrule
\end{tabularx}

\caption{A \textsc{TIDE-Bench} Joint example. Both models resolve $L1$ but miss the dependent record scope at $L2$; the resulting error persists after a correctly handled drift. SQL and model turns are abridged for space and clarity.}

\label{tab:case-study}
\vspace{-5mm}
\end{table}

\namedpara{Resolution across subtypes.}
To localize this gap, we decompose Drift-only DRA by drift subtype. Figure~\ref{fig:subtype} shows that resolution difficulty is not uniform across the three subtypes defined in \S~\ref{subsec:drift}: a consistent ordering emerges across models where projection-level drift is the easiest to resolve for 11 of 12 models and constraint-level drift is the hardest for 10 of 12, indicating that the resolution gap is structured rather than evenly distributed. The spread is also substantial: within a single model, the subtype-level DRA can differ by as much as 19.4 percentage points (GPT-5.4: projection 61.7\%, constraint 42.3\%). This ordering runs against what SQL complexity alone would predict. Constraint-level drift yields the \emph{simplest} post-drift SQL among the three subtypes, removing only WHERE predicates while leaving join, aggregation, and subquery structure unchanged (Appendix~\ref{app:complexity}). \textbf{\textit{The subtype that yields the simplest SQL is the hardest to resolve, indicating that what makes constraint-level drift difficult is not the SQL itself but inferring which prior commitment is now being retracted}}.

\subsection{How Chain and Drift Interact}
\label{subsec:interaction-finding}

What happens when chain and drift co-occur? Comparing Drift-only to Joint isolates the cost of placing the same drift after a chain phase, with the post-drift ground-truth SQL held identical across the two conditions by construction (\S~\ref{subsec:paired-factorial}). Figure~\ref{fig:slope} traces the shift from Drift-only to Joint across metrics. \textbf{\textit{The drop is asymmetric across all 12 models: DRR shifts by at most 5 percentage points, DRA falls by 6--12, and DEA by 13--18}}. Recognition depends only on the drift turn and is largely unaffected by upstream events; resolution and execution depend on the agent state when the drift arrives. If chain and drift failed independently, Joint DEA would equal the product of Chain-only and Drift-only accuracies, approximately 13\%. Across models, Joint DEA exceeds this prediction by 10--13 percentage points, \textbf{\textit{indicating that the samples on which the two failures occur largely overlap}}.

Across all three analyses, chain ambiguity, user intent drift, and their interaction emerge as three distinct dialogue-level failure modes that current models do not yet handle consistently and reliably.

\subsection{Case Study}
\label{subsec:case-study}

To make the interaction between chain ambiguity and user intent drift concrete, Table~\ref{tab:case-study} traces a Joint anchor through its chain and drift phases. In the chain phase, the models must determine whether a \emph{disappointing} season means an overall losing record or poor away-game performance, and then decide whether the record should reflect the team's full season or the coach's stint. Both models identify the first layer but fail to surface the dependent record-scope ambiguity, applying \texttt{coaches.won/lost} instead of the gold \texttt{teams.won/lost}. In the drift phase, the user removes the 2000--2010 restriction. Both models correctly delete the year predicate (DRR=True), yet retain the unresolved record-scope error, yielding preEX=False and postEX=False. Because this constraint-removal drift leaves the record definition unchanged, the post-drift failure can be cleanly traced to the chain phase. This example shows that resolving $L1$ and correctly applying the drift are not sufficient for end-to-end correctness: the unresolved $L2$ ambiguity invalidates both queries.
\section{Related Work}
\label{sec:related}

\noindent\textbf{Static benchmarks.}
Text-to-SQL progress has been driven by single-turn benchmarks. Spider~\citep{yu2018spider} established cross-domain generalization; BIRD~\citep{li2023bird} scaled evaluation to messy, knowledge-grounded databases across professional domains; and Spider~2.0~\citep{lei2025spider2} pushes into enterprise workflows with multiple SQL dialects. These benchmarks advance query generation under the assumption of a well-formed question, leaving multi-turn interaction untested.

\namedpara{Conversational and multi-turn benchmarks.}
Related work brings dialogue into text-to-SQL evaluation. SParC~\citep{yu2019sparc} and CoSQL~\citep{yu2019cosql} introduce context-dependent question sequences and Wizard-of-Oz dialogues over Spider databases. More recent benchmarks raise interactive realism: BIRD-Interact~\citep{huo2026interact} couples each database with a function-driven user simulator and a CRUD-spanning task suite, and DySQL-Bench~\citep{sun2025dysql} evaluates dynamic refinement under evolving user intent. These benchmarks mainly measure end-to-end multi-turn execution, but none specifically targets the \emph{conditional dependency} between clarification layers or the \emph{non-monotonic retraction} of committed elements as evaluation dimensions. \textsc{TIDE-Bench} addresses these two phenomena directly, with a paired factorial design that attributes performance differences to chain ambiguity and user intent drift.

\namedpara{Ambiguity and clarification.}
Prior work treats ambiguity as central. AmbiQT~\citep{bhaskar2023ambiqt} characterizes lexical and structural ambiguity from overlapping schema names and join paths; AMBROSIA~\citep{saparina2024ambrosia} grounds three linguistic ambiguity types in controlled databases; PRACTIQ~\citep{dong2025practiq} pairs ambiguous and unanswerable questions with single-turn clarifications; and AmbiSQL~\citep{ding2026ambisql} couples a fine-grained taxonomy with interactive multiple-choice resolution. HiL-Bench~\citep{trinh2026hilbench} studies help-seeking across progressively discovered but independent blockers. These ambiguity-focused benchmarks define rich turn-local taxonomies for individual user utterances. \textsc{TIDE-Bench} is complementary: it studies how ambiguities depend on one another across turns (chain) and how committed intent can be retracted across turns (drift), two dialogue-level phenomena beyond these turn-local taxonomies.

\section{Conclusion}
\label{sec:conclusion}

We introduce \textsc{TIDE-Bench}, a benchmark for evaluating how text-to-SQL agents handle two underexamined  patterns in conversational interaction: chain ambiguity and user intent drift. Built around a paired factorial design over 514 anchor SQLs from BIRD, \textsc{TIDE-Bench} contains 1{,}542 samples constructed through a four-stage automated pipeline with multi-stage validation and human verification. Our experiments on 12 LLMs reveal three failure modes. Chain ambiguity exposes an identification bottleneck unaffected by ask behavior, while user intent drift reveals a wide gap between recognition and resolution, and the two failures overlap when they co-occur. The paired design attributes these gaps to interaction structure rather than uncontrolled differences in schemas or query difficulty. These findings suggest that future work on conversational text-to-SQL should look beyond surface clarification behaviors toward modeling identification capacity and tracking dialogue-level commitments. This requires agents to surface dependent ambiguities and revise prior commitments without inheriting earlier errors. We hope \textsc{TIDE-Bench} provides a diagnostic tool for both research and development in conversational text-to-SQL.
\section*{Limitations}
\label{sec:limitations}

This study has limitations. First, although BIRD offers cross-domain coverage, generalization to other benchmarks remains untested. Second, chain ambiguity is restricted to two clarification layers; extending the counterfactual test to deeper chains remains future work. Third, Joint samples follow only a chain-then-drift order, leaving drift-then-chain unexplored. Finally, our evaluation relies on an LLM-based user simulator. Although the simulator is validated on \textsc{UserSim-Guard} and on expert-labeled traces (Fleiss' $\kappa = 0.802$), it remains an imperfect proxy; human-in-the-loop evaluation would better reflect real user behavior across practical settings and more clearly reveal simulator-specific biases.

\section*{Acknowledgments}
\label{sec:acknowledgments}

This work was supported in part by the National Natural Science Foundation of China (No. 62272200), and in part by the National Natural Science Foundation of China (No. 62502008), and in part by the Funding Scheme for Research and Innovation of FDCT (No. 0021/2025/ITP1), China and Macao S.A.R.
\nocite{*}
\bibliography{custom}

\clearpage
\appendix

\section{Commitment Extraction Rules}
\label{app:commitment-extractor}

\noindent\textbf{Overview.}
For each candidate query $y$, we define its commitment set $S(y)$ as a finite set of clause-scoped atomic units extracted from the parsed SQL abstract syntax tree. Each unit records a localized semantic decision expressed by the query, covering what is projected, which predicates are imposed, which grouping keys are used, and how the results are ordered. We obtain these units by traversing the abstract syntax tree produced by \texttt{sqlglot} and applying the per-scope emission rules in Table~\ref{tab:atomic-units}.

\namedpara{Extraction procedure.}
Algorithm~\ref{alg:commitment-extraction} constructs $S(y)$ using a scope-aware traversal. Each worklist entry pairs a query block with its scope path. Clause-level units are emitted only from the current scope, while immediate subqueries and set-operation branches are assigned distinct child paths and processed independently through the worklist.

\begin{table*}[t]
\small
\setlength{\tabcolsep}{6pt}
\renewcommand{\arraystretch}{1.45}
\begin{tabularx}{\linewidth}{@{}p{0.14\linewidth} X p{0.37\linewidth}@{}}
\toprule
\textbf{Scope} & \textbf{Emission rule} & \textbf{Example} \\
\midrule
\texttt{WHERE}
& Emit each conjunct after top-level \texttt{AND} decomposition; keep predicates
  under \texttt{OR} or \texttt{NOT} intact.
& \sql{WHERE a=1 AND b=2} $\to$ \{\sql{a=1}, \sql{b=2}\} \\

\texttt{SELECT}
& Emit each \texttt{SELECT} item; \texttt{SELECT *} is one marker.
& \sql{SELECT c, AVG(v)} $\to$ \{\sql{c}, \sql{AVG(v)}\} \\

\texttt{GROUP BY}
& Emit each grouping expression.
& \sql{GROUP BY c, d} $\to$ \{\sql{c}, \sql{d}\} \\

\texttt{ORDER BY}
& Emit each ordering expression with its direction.
& \sql{ORDER BY v DESC} $\to$ \{\sql{v DESC}\} \\

\texttt{HAVING}
& Emit each conjunct after top-level \texttt{AND} decomposition.
& \sql{HAVING COUNT(*)>5} $\to$ \{\sql{COUNT(*)>5}\} \\

\texttt{LIMIT}/\texttt{OFFSET}
& Emit the clause with its numeric argument.
& \sql{LIMIT 10} $\to$ \{\sql{LIMIT 10}\} \\
\bottomrule
\end{tabularx}
\vspace{-2.5mm}
\caption{Scope-specific emission rules for constructing the commitment set $S(y)$. Each emitted unit is paired with its query-scope path and canonicalized into value-preserving and structural forms before insertion for drift testing.}
\label{tab:atomic-units}
\vspace{-3mm}
\end{table*}

\namedpara{Value and structural forms.}
Each emitted unit is stored in two forms. The value form preserves literal values, such as \sql{region = 'West'}, while the structural form replaces literals with placeholders, such as \sql{region = ?}. The non-monotonicity test in \S\ref{subsec:drift} compares value forms: a follow-up is marked as drift when $S(y_{\mathrm{pre}}) \nsubseteq S(y_{\mathrm{post}})$. We use the structural form only to identify value-level edits: a retracted unit and a newly introduced unit share the same structure but differ only in their literal values.

\begin{algorithm}[t]
\captionsetup{
    font=footnotesize,
    labelfont=bf,
    textfont=normalfont
}
\caption{Commitment Extraction}
\label{alg:commitment-extraction}
\small
\renewcommand{\arraystretch}{1.08}

\begin{tabularx}{\linewidth}{
    @{}r@{\hspace{0.75em}}
    >{\raggedright\arraybackslash}X@{}
}

\multicolumn{2}{@{}l}{
    \textbf{Require:} Candidate SQL query $y$
} \\

\multicolumn{2}{@{}l}{
    \textbf{Ensure:} Commitment set $S(y)$
} \\[2pt]

\multicolumn{2}{@{}l}{
    \textbf{Notation:}
} \\[-1pt]

&
$W$: worklist of query-scope pairs $(q,\pi)$
\\

&
$\epsilon$: root scope
\\

&
$\pi\oplus\ell$: extend path $\pi$ with location $\ell$
\\

&
$\textsc{Branch}(q,j)$: path label for branch $b_j$
\\[2pt]

&
\textcolor{gray}{\itshape\% Parse and initialize}
\\[-1pt]

1: &
$T\gets\textsc{ParseSQL}(y)$;
$S\gets\emptyset$;
$W\gets[(T,\epsilon)]$
\\[2pt]

&
\textcolor{gray}{\itshape\% Traverse query scopes}
\\[-1pt]

2: &
\textbf{while} $W\neq\emptyset$ \textbf{do}
\\

3: &
\hspace*{1.2em}
$(q,\pi)\gets\textsc{Pop}(W)$
\\

4: &
\hspace*{1.2em}
\textbf{if} $q$ is a set operation \textbf{then}
\\

5: &
\hspace*{2.4em}
\textbf{for each} ordered branch $b_j$ of $q$ \textbf{do}
\\

6: &
\hspace*{3.6em}
$\pi_j\gets\pi\oplus\textsc{Branch}(q,j)$;
$\textsc{Push}(W,(b_j,\pi_j))$
\\

7: &
\hspace*{2.4em}
\textbf{end for}
\\

8: &
\hspace*{2.4em}
\textbf{continue}
\\

9: &
\hspace*{1.2em}
\textbf{end if}
\\[2pt]

&
\textcolor{gray}{\itshape\% Emit current-scope commitments}
\\[-1pt]

10: &
\hspace*{1.2em}
\textbf{for each} current-scope clause node $v$ of $q$
\textbf{do}
\\

11: &
\hspace*{2.4em}
\textbf{for each} unit $u$ emitted from $v$ by
Table~\ref{tab:atomic-units} \textbf{do}
\\

12: &
\hspace*{3.6em}
$u^{\mathrm{val}}\gets\textsc{CanonValue}(u)$;
$u^{\mathrm{str}}\gets\textsc{CanonStruct}(u)$
\\

13: &
\hspace*{3.6em}
$S\gets S\cup
\{(\pi,u^{\mathrm{val}},u^{\mathrm{str}})\}$
\\

14: &
\hspace*{2.4em}
\textbf{end for}
\\

15: &
\hspace*{1.2em}
\textbf{end for}
\\[2pt]

&
\textcolor{gray}{\itshape\% Enqueue nested subqueries}
\\[-1pt]

16: &
\hspace*{1.2em}
\textbf{for each} immediate subquery $q_j$ at $\ell_j$ in $q$
\textbf{do}
\\

17: &
\hspace*{2.4em}
$\pi_j\gets\pi\oplus\ell_j$;
$\textsc{Push}(W,(q_j,\pi_j))$
\\

18: &
\hspace*{1.2em}
\textbf{end for}
\\

19: &
\textbf{end while}
\\[2pt]

20: &
\textbf{return} $S$
\\

\end{tabularx}

\end{algorithm}

\section{Construction Pipeline Details}
\label{app:pipeline}

\begin{table}[t]
\centering
\small
\setlength{\tabcolsep}{4.5pt}
\renewcommand{\arraystretch}{1.50}
\begin{tabularx}{\linewidth}{@{}p{0.25\linewidth} X@{}}
\toprule
\textbf{Criterion} & \textbf{Requirement} \\
\midrule
Plausibility
& L1 and L2 options are plausible and distinct, with no strawmen or rewordings. \\
Re-ask
& Each layer adds a dimension absent from the initial question. \\
Counterfactuality
& L1 must eliminate or transform the core of L2, rather than leave it invariant. \\
\bottomrule
\end{tabularx}
\vspace{-2mm}
\caption{LLM judging rubric for chain injection. Both judges must pass three criteria for benchmark inclusion.}
\label{tab:chain-judge-rubric}
\vspace{-6mm}
\end{table}

This appendix expands the construction pipeline summarized in \S\ref{sec:construction}. For added clarity, We detail anchor selection, chain injection, drift injection, and evidence rewriting in order, specifying how candidates are selected, generated, filtered, and verified before their inclusion in \textsc{TIDE-Bench}.

\begin{table*}[t]
\centering
\small
\setlength{\tabcolsep}{5pt}
\renewcommand{\arraystretch}{1.60}
\begin{tabularx}{\linewidth}{@{}c p{0.15\linewidth} X p{0.22\linewidth}@{}}
\toprule
\textbf{Step} & \textbf{Check} & \textbf{Pass condition} & \textbf{Failure signal} \\
\midrule
1 & Output schema
& Required JSON fields are present.
& \texttt{json\_format} \\
2 & SQL parsing
& \texttt{post\_drift\_sql} parses and yields extractable commitments.
& \texttt{sql\_parse} \\
3 & SQL execution
& \texttt{post\_drift\_sql} executes on the target database.
& \texttt{sql\_execute} \\
4 & Non-empty result
& The post-drift query returns at least one row.
& \texttt{empty\_result} \\
5 & Observable effect
& Pre- and post-drift result sets differ.
& \texttt{identical\_result} \\
6 & Non-monotonicity
& The commitment diff satisfies $S_{\mathrm{pre}} \nsubseteq S_{\mathrm{post}}$.
& \texttt{no\_change}; \texttt{monotonic} \\
7 & Subtype match
& The realized drift subtype matches the assigned subtype.
& \texttt{subtype\_mismatch} \\
8 & Subtype detail
& For projection-level samples, the fine-grained subtype matches.
& \texttt{fine\_subtype} \\
\bottomrule
\end{tabularx}
\vspace{-0.5mm}
\caption{Sequential AST-level checks for validating drift candidates. Checks are applied in order; the final column reports the rejection signal emitted upon failure, and only candidates passing all checks undergo LLM judging.}
\label{tab:drift-ast}
\vspace{-4mm}
\end{table*}

\begin{table}[t]
\centering
\small
\setlength{\tabcolsep}{4.5pt}
\renewcommand{\arraystretch}{1.60}
\begin{tabular}{@{}cccccc@{}}
\toprule
Cond. & Evidence & Question & Dialogue & Acc. & $\Delta$\\
\midrule
\textbf{A} & orig. & orig.     & ---        & 51.2 & ---    \\
\textbf{B} & rew.  & ambiguous & \checkmark & 50.4 & $-0.8$ \\
\textbf{C} & rew.  & ambiguous & ---        & 31.2 & $-20.0$ \\
\bottomrule
\end{tabular}
\vspace{-0.5mm}
\caption{Reconstructability of rewritten evidence under three input settings ($n=695$). ``orig.''~/~``rew.'' denote original~/~rewritten BIRD evidence; \checkmark marks dialogue input; $\Delta$ is the overall accuracy change (pp) from A.}
\label{tab:evidence-ablation}
\vspace{-5mm}
\end{table}

\subsection{Chain Injection}
\label{app:pipeline-chain}
\noindent\textbf{Generation.}
We use GPT-5.1 to generate the chain candidates. For each BIRD entry, the model produces a JSON object with the initial question and the L1/L2 clarification pairs. The full prompt template and output schema are both shown in Figure~\ref{fig:chain-prompt}.

\namedpara{Judging.}
We score each candidate with two LLM judges, GPT-5.3-Codex and DeepSeek-V4-Pro. Each judge applies the three criteria in Table~\ref{tab:chain-judge-rubric}, in priority order. A candidate is accepted only if both judges pass all three criteria. The judges agree on 94.7\%, and 695/830 candidates pass (83.7\%).

\subsection{Drift Injection}
\label{app:pipeline-drift}

\noindent\textbf{Generation.}
We use GPT-5.1 to generate drift candidates. For each chain-validated anchor, the model receives a drift subtype: value-, constraint-, or projection-level. It returns a JSON object with the drift utterance and post-drift SQL. The prompt template and output schema are shown in Figure~\ref{fig:drift-prompt}.

\namedpara{AST-level validation.}
Each candidate is checked by a rule-based validator before LLM judging. The validator tests the operational definition of \S\ref{subsec:drift} and the scope constraints of the assigned subtype, as summarized in Table~\ref{tab:drift-ast}. The validation reuses the AST procedure from Appendix~\ref{app:commitment-extractor}; 554 of 695 validated chain candidates pass this stage (79.7\%).

\namedpara{Cross-validation.}
AST-passing candidates are scored by two LLM judges, Claude-Sonnet-4.6 and DeepSeek-V4-Pro. The judges receive the anchor SQL, drift utterance, and post-drift SQL. They apply two criteria in priority order: (i) \textbf{Drift fit}, requiring the post-drift SQL to match the assigned subtype and user drift; and (ii) \textbf{Stability}, requiring untouched units to remain unchanged. A candidate is accepted only if both judges pass both criteria, requiring agreement on validity. The judges agree on 96.6\%, and 514/554 candidates pass (92.8\%).

\subsection{Evidence Rewriting}
\label{app:pipeline-evidence}

\noindent\textbf{Motivation.}
BIRD's \texttt{evidence} field can reveal terms that should stay ambiguous in \textsc{TIDE-Bench}. For example, it may map ``top clients'' to ``clients ranked by spend,'' fixing the L1 choice ahead of dialogue. This early cue weakens the chain ambiguity we aim to test. The rewrite removes such cues while preserving schema-essential mappings.

\namedpara{Protocol.}
For each sample, GPT-5.1 rewrites the evidence using the source evidence, anchor SQL, and generated L1/L2 clarifications. The rewrite removes cues that resolve the ambiguity while retaining mappings needed for SQL interpretation.

\namedpara{Reconstructability.}
We test whether the rewritten evidence preserves schema-essential information without itself revealing the L1/L2 resolution. For 695 candidates, we evaluate Gemini-3-Flash under three settings in Table~\ref{tab:evidence-ablation}: the original question and evidence (A), the ambiguous question and rewritten evidence with dialogue (B), and the same inputs without dialogue (C). The small A--B gap of 0.8 pp shows that the rewritten evidence preserves what is needed for reconstruction when paired with dialogue. The large B--C gap of 19.2 pp shows that the rewritten evidence alone does not leak resolution.

\section{Human Validation Details}
\label{app:validation}

This appendix supplements \S\ref{sec:human-validation} with the annotator protocol, the full checklist, the adjudication rules, and the raw counts behind the item-level statistics.

\subsection{Annotator Recruitment and Protocol}
\label{app:validation-annotators}

The audit is performed by five annotators: three NLP PhD students with text-to-SQL research experience, and two recruited annotators with production SQL backgrounds. The two external annotators are compensated at hourly rates commensurate with local expectations for technical annotation work. Before annotating the audit set, all five annotators completed a one-hour calibration round on a held-out set of ten triplets, drawn independently from the audit sample. The calibration
round aligned checklist-item interpretation; calibration triplets are excluded from all reported statistics.

Each of the 250 audit triplets is assigned to all five annotators, and each annotator works independently. Annotators see only the sample contents---initial question, clarification structure, drift utterance (if applicable), and ground-truth SQL---and are blind to all automated verdicts produced during construction, as well as to any model evaluation results. The annotation interface presents one sample at a time, with the checklist items relevant to that sample's type. Items that are not applicable to a given sample type (e.g., joint-boundary items on a \textsc{Chain-only} sample) are simply not shown there.

\subsection{Full Checklist}
\label{app:validation-checklist}

Each annotator answers yes/no on the following items. The four chain criteria apply to \textsc{Chain-only} and \textsc{Joint} samples; the three drift criteria apply to \textsc{Drift-only} and \textsc{Joint} samples; the joint criterion applies only to \textsc{Joint} samples, where both patterns co-occur within the same dialogue.

\paragraph{Chain items.}

\begin{itemize}[
  topsep=0.15em,
  itemsep=0.05em,
  parsep=0pt,
  partopsep=0pt
]
\item[\textbf{1.}] Is the initial user utterance genuinely underspecified with respect to the business concept clarified at $L_1$?
\item[\textbf{2.}] Are both $L_1$ options plausible interpretations of that concept that a real user might intend?
\item[\textbf{3.}] Are both $L_2$ options plausible interpretations of the concept clarified at $L_2$?
\item[\textbf{4.}] Under the $L_1$ alternative, is the core of $L_2$ eliminated or transformed, rather than simply remaining unchanged in that context?
\end{itemize}

\vspace{-0.6mm}

\paragraph{Drift items.}
\begin{itemize}[
  topsep=0.15em,
  itemsep=0.05em,
  parsep=0pt,
  partopsep=0pt
]
\item[\textbf{1.}] Does the drift utterance retract or replace at least one commitment from the pre-drift turn, rather than only extending it?
\item[\textbf{2.}] Does the post-drift SQL correctly implement what the drift utterance asks for?
\item[\textbf{3.}] Does the post-drift SQL leave every commitment not targeted by the drift unchanged?
\end{itemize}

\vspace{-0.6mm}

\paragraph{Joint item.}
\begin{itemize}[
  topsep=0.15em,
  itemsep=0.05em,
  parsep=0pt,
  partopsep=0pt
]
\item[\textbf{1.}] Does the dialogue read coherently across the chain-to-drift boundary, with no abrupt topic shift and no contradiction between turns?
\end{itemize}

\subsection{Adjudication Rules}
\label{app:validation-adjudication}

A sample is accepted under \emph{majority acceptance} iff at least three of the five annotators answer yes on every applicable item; it is accepted under \emph{strict acceptance} iff all five annotators answer yes on every applicable item. Items that are not applicable to a sample's type are excluded from both the numerator and the denominator for that sample only.

Fleiss' $\kappa$ in Table~\ref{tab:human-validation} is computed over binary per-annotator accept/reject verdict at the sample level, where a verdict is yes iff that annotator answered yes on every applicable item. The item-level $\kappa$ values below are computed analogously over binary per-annotator verdicts for every checklist item.

\subsection{Item-Level Statistics}
\label{app:validation-item-level}

Table~\ref{tab:human-validation} reports sample-level acceptance after aggregating all applicable checklist criteria. To identify where rejection and disagreement arise, Table~\ref{tab:item-level-app} reports majority pass rates and Fleiss' $\kappa$ for three core criteria: counterfactual dependency, non-monotonic retraction, and joint transition coherence. A criterion passes for a given sample iff at least three of the five annotators mark it as satisfied.

All three item-level $\kappa$ values lie between $0.71$ and $0.83$, indicating substantial to almost-perfect agreement. The lowest value ($\kappa = 0.71$, on joint transition coherence) is consistent with the more pragmatic nature of that judgment, while the two formal criteria show higher agreement overall.

\begin{table}[!t]
\centering
\small
\setlength{\tabcolsep}{5pt}
\renewcommand{\arraystretch}{1.45}

\begin{tabular*}{\linewidth}{@{\extracolsep{\fill}}ccc@{}}
\toprule
\textbf{Item}
& \textbf{Pass Rate}
& \textbf{Fleiss' $\boldsymbol{\kappa}$} \\
\midrule
Counterfactual dependency
& 93.4\%
& 0.79 \\

Non-monotonic retraction
& 94.8\%
& 0.83 \\

Joint transition coherence
& 91.6\%
& 0.71 \\
\bottomrule
\end{tabular*}
\vspace{-2mm}
\caption{Majority pass rates and annotator agreement for three core validation criteria. A criterion passes when at least three of five annotators mark it as satisfied; Fleiss' $\kappa$ uses binary criterion-level verdicts across annotators.}
\label{tab:item-level-app}
\vspace{-3mm}
\end{table}

\subsection{Dominant Failure Modes}
\label{app:validation-failures}

We summarize the dominant reasons for rejection under majority acceptance for each sample type:

\namedpara{Chain-only.}
Of the 17 rejected samples, the most common failure concerns counterfactual dependency. The remaining rejections involve the plausibility of the $L_1$ or $L_2$ options, while initial under-specification rarely emerges as a rejection reason.

\namedpara{Drift-only.}
Of the 11 rejected samples, the most common failure concerns post-drift SQL fidelity to the drift utterance. Fewer failures involve the preservation of non-targeted commitments, while non-monotonic retraction itself is rarely flagged.

\namedpara{Joint.}
The 29 rejected samples most often fail on counterfactual dependency, non-monotonic retraction, or chain-to-drift boundary coherence. The remaining rejections reflect less frequent failures on other checklist items across the full audit set.
\section{Evaluation Setup}
\label{app:evaluation-setup}

\noindent\textbf{Models.} We evaluate 12 frontier LLMs, all accessed via official APIs: Claude-Sonnet-4.6, GPT-4o, GPT-5.4, o3-mini, o4-mini, Gemini-2.5-Flash, Grok-4-Fast, Grok-4.1-Fast, Qwen2.5-72B, Qwen-Max, Llama-3.3-70B, and DeepSeek-V4-Flash. Both reasoning and
non-reasoning models are included to span the current capability frontier; exact model configurations are also listed in Table~\ref{tab:model_configs}.

\namedpara{Dialogue harness.} Each evaluation runs an agent--simulator dialogue loop bounded by a maximum of 20 turns. Each agent turn yields either a clarification request or a candidate SQL query, parsed from a JSON response under the schema described below. Malformed responses trigger a single retry; persistent failures count as an abstention and the trajectory terminates. Candidate SQL is executed against the corresponding BIRD SQLite database under read-only mode; execution results are not surfaced back to the agent at any subsequent turn.

\namedpara{Inputs to the agent.} At each turn, the agent receives the database DDL, the rewritten
evidence (\S\ref{subsec:pipeline}, with clauses pre-resolving L1 or L2 removed), and the dialogue history through that turn. The agent prompt is fixed across all 12 models.

\namedpara{Agent prompt.}
The system prompt template is shown in
Figure~\ref{fig:agent-prompt}. It defines the required JSON response
schema and the permitted agent actions.

\begin{table}[t]
\centering
\small
\renewcommand{\arraystretch}{1.30}
\begin{tabular}{ll}
\toprule
\textbf{Display name} & \textbf{Model configuration} \\
\midrule
Claude Sonnet 4.6 & \texttt{claude-sonnet-4.6} \\
Qwen2.5-72B & \texttt{qwen2.5-72b-instruct} \\
o4-mini & \texttt{o4-mini} \\
GPT-4o & \texttt{gpt-4o-2024-11-20} \\
GPT-5.4 & \texttt{gpt-5.4} \\
Llama-3.3-70B & \texttt{llama-3.3-70b-instruct} \\
o3-mini & \texttt{o3-mini} \\
Gemini-2.5-Flash & \texttt{gemini-2.5-flash-nothinking} \\
Grok-4.1-Fast & \texttt{grok-4-1-fast-reasoning} \\
Qwen-Max & \texttt{qwen-max} \\
DeepSeek-V4-Flash & \texttt{deepseek-chat} \\
Grok-4-Fast & \texttt{grok-4-fast-non-reasoning} \\
\bottomrule
\end{tabular}
\caption{Display names and API model configurations for the 12
evaluated LLMs. The configuration strings identify exact model
variants and, where applicable, their reasoning modes used in all
reported evaluations.}
\label{tab:model_configs}
\vspace{-4mm}
\end{table}
\section{User Simulator Details}
\label{app:user-sim}

\noindent\textbf{Two-stage design.} The simulator follows the function-driven design of  \citet{huo2026interact}. At each user turn responding to an agent  clarification, a Stage~1 router classifies the request into one of  three actions: \textsc{Amb} (resolve a pre-annotated ambiguity from  the sample's chain structure), \textsc{Loc} (answer a localizable SQL-level detail by inspecting the ground-truth SQL), or \textsc{Una} (refuse a request that the user cannot reasonably be expected to answer, such as questions about database-internal identifiers). Stage~2 then generates the natural-language response conditioned on the routed action and the relevant annotation field  for that sample.

\namedpara{Adaptations for \textsc{TIDE-Bench}.} Three changes accommodate the two-layer chain structure 
(\S\ref{subsec:chain}):
\begin{itemize}[leftmargin=1.2em,itemsep=0.18em,topsep=0.25em]
\item L2-targeted clarifications issued while L1 is still unresolved are routed as \textsc{Loc} rather than \textsc{Amb}, since the L2 annotation only becomes a valid \textsc{Amb} target after L1 has been explicitly chosen.

\item AST traversal during Stage~2 generation is restricted to a single atomic node per response, preventing the simulator from volunteering information beyond what the agent asked for.

\item A post-generation validator screens responses for out-of-scope SQL fragments or schema identifiers; offending responses are regenerated.
\end{itemize}

\namedpara{Configuration.}
Both Stage~1 routing and Stage~2 generation use GPT-5.1 via API, with the same temperature and decoding settings as the agent harness (Appendix~\ref{app:evaluation-setup}) to ensure configuration consistency.

\namedpara{Validation.}
We evaluate the simulator on the USERSIM-GUARD diagnostic suite, which probes the three router actions on held-out queries. Per-split accuracy is reported in Table~\ref{tab:usersim-guard}. To verify behavior in full dialogues, the five annotators from \S\ref{sec:human-validation} additionally labeled a stratified random sample of agent--simulator traces (Fleiss' $\kappa = 0.802$).

\begin{table}[!t]
\centering
\small
\renewcommand{\arraystretch}{1.25}

\begin{tabular*}{\linewidth}{
@{\hspace{12pt}}
c
@{\extracolsep{\fill}}
c
@{\hspace{12pt}}
}
\toprule
\textbf{Split}
& \textbf{Accuracy (\%)} \\
\midrule
\textsc{Una} (refusal)   & 97.4 \\
\textsc{Amb} (labeled)   & 94.3 \\
\textsc{Loc} (unlabeled) & 93.0 \\
\bottomrule
\end{tabular*}
\vspace{-1mm}
\caption{Stage~1 router accuracy on the three \textsc{UserSim-Guard} splits. \textsc{Una} tests refusal decisions, \textsc{Amb} tests labeled ambiguity routing, and \textsc{Loc} tests unlabeled SQL-detail localization during interaction.}
\label{tab:usersim-guard}
\vspace{-5mm}
\end{table}
\section{Full Per-Model Results}
\label{app:chain-id}

Tables~\ref{tab:chain-identification} and~\ref{app:drift-handling} provide the complete per-model breakdowns behind the summary in \S\ref{subsec:performance}. Table~\ref{tab:chain-identification} reports chain-layer identification
on Chain-only and Joint samples, together with ask rates across Chain-only, Joint, and Drift-only settings. Table~\ref{app:drift-handling} reports drift recognition and resolution on Drift-only and Joint
samples, broken down by drift subtype. Two patterns hold across all twelve models.
\section{Ground-Truth SQL Complexity}
\label{app:complexity}

This appendix substantiates two claims made in \S\ref{sec:experiment}: (a) value-level drift produces post-drift SQL that is structurally identical to the pre-drift query (\S\ref{subsec:interaction-finding}), and (b) constraint-level drift produces the simplest post-drift SQL among the three subtypes (\S\ref{subsec:drift-bottleneck}). We characterize each drift subtype by comparing the pre-drift ground-truth SQL $y_{\mathrm{pre}}^*$ against the post-drift ground-truth SQL $y_{\mathrm{post}}^*$ over all 514 triplets. All anchors parse successfully under our analyzer.

\subsection{Per-Subtype Structural Effects}
\label{subapp:subtype-effects}

Table~\ref{tab:complexity-subtype} reports the mean change ($\Delta$) in five key structural complexity measures from pre- to post-drift, separately for each drift subtype. Values are means over the corresponding subtype subset; negative values indicate that the post-drift SQL is
simpler than the pre-drift SQL along that dimension.

\namedpara{Value-level.} Every measure shows zero mean change. This is by design: value-level drift only replaces a literal inside a WHERE predicate, with predicate structure and all other commitments preserved (\S\ref{subsec:drift}). The AST validator explicitly enforces this invariant for every candidate during construction, and the tier distribution before and after drift is identical (44/107/24 in basic/intermediate/advanced). For value-level samples, $y_{\mathrm{pre}}^*$ and $y_{\mathrm{post}}^*$ are therefore structurally indistinguishable at the level of these measures, isolating dialogue state itself as the only variable that still differs between the paired Drift-only and Joint samples (\S\ref{subsec:interaction-finding}).

\namedpara{Constraint-level.} WHERE predicates drop by 1.11 on average across constraint-level samples, consistent with the subtype definition (retraction of one or more WHERE predicates without introducing new constraints, \S\ref{subsec:drift}). Token count drops by 7.47, the largest absolute reduction among the three subtypes. Critically, JOIN, aggregation, and subquery structures are all unchanged on average. The tier distribution shifts substantially toward lower complexity (44/104/13 $\to$ 76/79/6 in basic/intermediate/advanced), with roughly 25 samples moving from intermediate to basic. By every measure reported, constraint-level drift yields the simplest post-drift SQL of the three subtypes (used in \S\ref{subsec:drift-bottleneck}).

\begin{table}[t]
\centering
\small
\setlength{\tabcolsep}{4pt}
\renewcommand{\arraystretch}{1.45}
\begin{tabular}{lccc}
\toprule
& \textbf{Value} & \textbf{Constraint} & \textbf{Projection} \\
& ($n{=}175$) & ($n{=}161$) & ($n{=}178$) \\
\midrule
SQL Tokens         & $+0.00$ & $-7.47$ & $-5.87$ \\
WHERE Predicates   & $+0.00$ & $-1.11$ & $-0.03$ \\
JOIN Count         & $+0.00$ & $+0.00$ & $-0.03$ \\
Aggregation Count  & $+0.00$ & $+0.00$ & $-0.26$ \\
Subquery Depth     & $+0.00$ & $+0.00$ & $-0.01$ \\
\bottomrule
\end{tabular}
\vspace{-2mm}
\caption{Mean change in structural complexity from $y_{\mathrm{pre}}^*$ to $y_{\mathrm{post}}^*$, by drift subtype. Values are means over the corresponding subtype subset. Negative values indicate reduced post-drift complexity along the corresponding dimension, with zero denoting no change on average.}
\label{tab:complexity-subtype}
\vspace{-5mm}
\end{table}

\namedpara{Projection-level.} Aggregation count drops by 0.26 on average, while WHERE predicates ($-0.03$) and JOIN count ($-0.03$) are effectively unchanged. This matches the subtype definition: changes are confined to SELECT, GROUP BY, ORDER BY, HAVING, LIMIT, or OFFSET, with WHERE predicates preserved (\S\ref{subsec:drift}). The aggregation reduction reflects projection changes that move from aggregate to non-aggregate queries (e.g., from \texttt{SELECT AVG(profit)} to a top-$k$ row listing query), while largely preserving the filtering and join structure.

\begin{table*}[t]
\centering
\small
\setlength{\tabcolsep}{3pt}
\renewcommand{\arraystretch}{1.45}

\begin{tabular*}{\linewidth}{
@{}
c@{\hspace{4pt}}
l@{\hspace{6pt}}
c@{\extracolsep{\fill}}
c c c c c c c c
@{}
}
\toprule
\multirow{2}{*}{\textbf{\#}}
& \multicolumn{1}{c@{\hspace{6pt}}}
  {\multirow{2}{*}{\textbf{Model}}}
& \multicolumn{4}{c}{\textbf{Chain-only}}
& \multicolumn{4}{c}{\textbf{Joint}}
& \multicolumn{1}{c}{\textbf{Drift-only}} \\
\cmidrule(lr){3-6}
\cmidrule(lr){7-10}
\cmidrule(lr){11-11}

& &
\textbf{L1-IR}
& \textbf{L2-IR$^p$}
& \textbf{JIR}
& \textbf{AR}
& \textbf{L1-IR}
& \textbf{L2-IR$^p$}
& \textbf{JIR}
& \textbf{AR}
& \textbf{AR} \\
\midrule
1  & Claude-Sonnet-4.6 & 5.6  & 13.8 & 0.6 & 9.7  & 6.0  & 11.1 & 1.4 & 10.5 & 0.4  \\
2  & Qwen2.5-72B       & 36.2 & 3.3  & 1.2 & 60.1 & 35.8 & 3.9  & 1.8 & 60.7 & 20.0 \\
3  & o4-mini           & 27.8 & 11.5 & 1.4 & 54.9 & 29.2 & 12.6 & 2.1 & 55.4 & 17.5 \\
4  & GPT-4o            & 31.3 & 5.3  & 1.4 & 48.2 & 30.0 & 6.2  & 4.1 & 47.3 & 8.9  \\
5  & GPT-5.4           & 19.5 & 11.7 & 2.5 & 30.0 & 20.0 & 10.9 & 3.1 & 31.7 & 2.5  \\
6  & Llama-3.3-70B     & 32.7 & 12.1 & 3.7 & 58.2 & 34.0 & 12.1 & 3.3 & 58.0 & 11.1 \\
7  & o3-mini           & 21.6 & 9.9  & 1.8 & 33.5 & 23.7 & 9.7  & 2.1 & 34.8 & 4.3  \\
8  & Gemini-2.5-Flash  & 15.6 & 1.2  & 1.2 & 30.5 & 12.6 & 0.4  & 0.4 & 28.8 & 4.1  \\
9  & Grok-4.1-Fast     & 26.1 & 12.6 & 2.9 & 39.5 & 25.7 & 11.5 & 2.1 & 42.2 & 5.4  \\
10 & Qwen-Max          & 27.8 & 1.4  & 1.4 & 43.8 & 26.5 & 2.1  & 2.1 & 44.9 & 4.3  \\
11 & DeepSeek-V4-Flash & 8.4  & 5.1  & 0.4 & 14.6 & 7.8  & 5.1  & 0.6 & 12.8 & 0.8  \\
12 & Grok-4-Fast       & 18.9 & 11.7 & 1.6 & 34.6 & 18.7 & 11.1 & 1.8 & 36.4 & 5.4  \\
\bottomrule
\end{tabular*}

\vspace{-1mm}
\caption{Chain-layer identification and ask rates (\%) on Chain-only,
Joint, and Drift-only samples. L1-IR and L2-IR$^p$ measure whether the
agent identifies the corresponding clarification layer; JIR requires
both layers to be identified within one trajectory, while AR records
whether the agent issues any clarification request. L2-IR$^p$ is
measured under the $p$-protocol with L1 pre-injected; all other metrics
use the $f$-protocol throughout all
evaluations.}
\label{tab:chain-identification}
\vspace{-1mm}
\end{table*}

\begin{table*}[t]
\centering
\small
\setlength{\tabcolsep}{3pt}
\renewcommand{\arraystretch}{1.45}

\begin{tabular*}{\linewidth}{
@{}
c@{\hspace{4pt}}
l@{\hspace{6pt}}
c@{\extracolsep{\fill}}
c c c c c c c c c
@{}
}
\toprule
\multirow{2}{*}{\textbf{\#}}
& \multicolumn{1}{c@{\hspace{6pt}}}
  {\multirow{2}{*}{\textbf{Model}}}
& \multicolumn{4}{c}{\textbf{Drift-only DRA}}
& \multicolumn{4}{c}{\textbf{Joint DRA}}
& \multicolumn{2}{c}{\textbf{DRR}} \\
\cmidrule(lr){3-6}
\cmidrule(lr){7-10}
\cmidrule(lr){11-12}

& &
\textbf{Value}
& \textbf{Constraint}
& \textbf{Projection}
& \textbf{All}
& \textbf{Value}
& \textbf{Constraint}
& \textbf{Projection}
& \textbf{All}
& \textbf{Drift}
& \textbf{Joint} \\
\midrule
1  & Claude-Sonnet-4.6 & \textbf{53.4} & \underline{48.5} & \textbf{64.0} & \textbf{55.4} & \underline{38.6} & 34.4 & \textbf{55.4} & \textbf{43.0} & \textbf{98.2} & \underline{96.9} \\
2  & Qwen2.5-72B       & 47.2 & 46.0 & 58.3 & 50.6 & 38.3 & 35.6 & 45.7 & 40.0 & \textbf{98.2} & 95.8 \\
3  & o4-mini           & 46.6 & 44.2 & 59.4 & 50.2 & \textbf{39.7} & 32.5 & 49.7 & 40.8 & 94.9 & 93.7 \\
4  & GPT-4o            & \underline{49.4} & 45.1 & \underline{62.3} & \underline{52.4} & 34.1 & 35.0 & 51.4 & 40.3 & 97.3 & \underline{96.9} \\
5  & GPT-5.4           & 48.3 & 42.3 & 61.7 & 51.0 & 35.8 & 34.4 & \underline{52.6} & \underline{41.1} & 97.3 & 94.7 \\
6  & Llama-3.3-70B     & 44.9 & 44.8 & 52.0 & 47.3 & 38.1 & \textbf{39.9} & 45.1 & \underline{41.1} & 97.4 & \textbf{97.1} \\
7  & o3-mini           & 44.3 & 42.3 & 55.4 & 47.5 & 35.2 & 30.1 & 51.4 & 39.1 & 95.7 & 94.9 \\
8  & Gemini-2.5-Flash  & 44.3 & \textbf{49.7} & 47.4 & 47.1 & 36.4 & \underline{39.3} & 45.7 & 40.5 & 97.4 & 95.7 \\
9  & Grok-4.1-Fast     & 44.3 & 42.9 & 58.3 & 48.6 & 33.5 & 30.1 & 46.3 & 36.8 & 96.5 & 92.0 \\
10 & Qwen-Max          & 46.0 & 47.9 & 54.3 & 49.4 & 34.7 & 35.6 & 41.1 & 37.2 & \underline{97.9} & 96.1 \\
11 & DeepSeek-V4-Flash & 45.5 & 43.6 & 54.3 & 47.9 & 29.0 & 32.5 & 46.9 & 36.2 & 97.3 & 96.7 \\
12 & Grok-4-Fast       & 40.9 & 37.4 & 57.7 & 45.5 & 29.0 & 28.8 & 47.4 & 35.2 & 96.4 & 93.9 \\
\bottomrule
\end{tabular*}

\vspace{-1mm}
\caption{Drift handling metrics (\%) under the paired factorial design.
The best and second-best results in each column are highlighted in bold
and underlined, respectively. Drift-only DRA and Joint DRA are reported
overall (All) and by drift subtype. The DRR columns report aggregate
Drift Recognition Rate on Drift-only and Joint samples. Higher values
indicate more reliable recognition and resolution of injected drift
across interaction settings.}
\label{app:drift-handling}
\vspace{-1mm}
\end{table*}

\begin{figure*}[t]
\centering
\begin{tcolorbox}[
  width=1.03\textwidth,
  enhanced,
  colback=white,
  colframe=black,
  boxrule=0.5pt,
  arc=0.8mm,
  left=1.6mm,
  right=1.6mm,
  top=1.3mm,
  bottom=1.3mm
]
\small
\setstretch{1.04}
\setlength{\parskip}{0.32em}

\noindent System prompt.

\noindent
You are an expert data annotator for a conversational Text-to-SQL benchmark thatstudies chained ambiguity. Your task is to construct CHAIN AMBIGUITY samples on top of the BIRD benchmark.

A chained ambiguity sample is a dialogue where:
\begin{itemize}[leftmargin=1.35em,itemsep=0.05em,topsep=0.05em]
    \item A user issues an initially ambiguous question.
    \item The system asks a first clarification (L1), whose options represent two
    genuinely plausible interpretations of an abstract business concept.
    \item The user selects one option, \texttt{L1\_chosen}.
    \item ONLY AFTER L1 is resolved does a second ambiguity (L2) become meaningful.
    L2 must be activated by the specific L1 answer, not be a standalone ambiguity.
    \item The user answers L2, and the system outputs the final SQL, which must
    match the provided BIRD ground-truth SQL exactly.
\end{itemize}

\vspace{0.35em}
\noindent Hard constraints.

\begin{enumerate}[leftmargin=1.8em,itemsep=0.22em,topsep=0.15em,label=C\arabic*.]
    \item L1 must be a binary choice.
    \texttt{L1\_options} contains exactly 2 items. Both options must represent
    plausible user intents: a real user could reasonably have meant either one.

    \item L2 must be a binary choice.
    \texttt{L2\_options} contains exactly 2 items. Both options must represent
    plausible user intents: a real user could reasonably have meant either one.

    \item Ground-truth path.
    \texttt{L1\_chosen} must be the option that, together with
    \texttt{L2\_chosen}, leads to the BIRD ground-truth SQL. The other L1 option
    must lead to a different but valid SQL.

    \item Conditional dependency.
    L2 must become meaningful only after \texttt{L1\_chosen} is selected. Strip
    any opening reference to \texttt{L1\_chosen} from \texttt{L2\_question}
    and check the remaining core. If the same core question could be asked under
    the L1 alternative with the same words, then L2 is independent, not a chain.
    Redesign.

    \item Content novelty.
    Each clarification layer must introduce information not already present in
    the initial question. If a layer only re-asks or narrows concrete concepts,
    entities, filters, named values, or constraints already stated by the user,
    redesign it to cover a genuinely new dimension.

    \item Literal placement.
    Category A literals are ambiguity-relevant literals that operationalize an
    abstract business term; they must surface through L1 or L2 clarification.
    Category B literals are context parameters, including LIMIT values, dates,
    named subjects, places, products, and arbitrary numeric query parameters;
    they must appear in the initial question. Category C literals are explicitly
    mapped by the BIRD evidence field and need not appear in the dialogue.

    \item Initial question.
    The initial question must contain: (a) all Category B literals from the BIRD
    SQL; (b) the abstract business term whose operationalization is ambiguous;
    and (c) no Category A literals. It must be ambiguous with respect to L1 only
    and should not pre-expose L2.

    \item Correspondence to BIRD SQL.
    The resolved dialogue must fully specify the BIRD ground-truth SQL. Every
    filter, LIMIT, aggregation, ordering, and join must trace back to either a
    Category B literal in the initial question, a Category A literal grounded in
    L1/L2, or a Category C evidence mapping.

    \item Naturalness.
    User utterances must use natural business language. Users never reference SQL
    column names, table names, or schema identifiers. Clarification must be at
    the business-concept level, not the schema-detail level.

    \item Alternative SQL feasibility.
    Both the L1 alternative and L2 alternative must correspond to SQL actually
    writable against the provided schema. Do not invent columns, tables, or
    relationships. For each alternative, verify that the referenced columns,
    tables, and joins exist in the schema.
\end{enumerate}

\vspace{0.35em}
\noindent User prompt template.

\noindent
Construct a CHAIN AMBIGUITY sample from a BIRD entry containing
\texttt{db\_id}, \texttt{schema\_ddl}, \texttt{sample\_rows},
\texttt{bird\_question}, \texttt{bird\_evidence}, \texttt{bird\_sql}, and
\texttt{complexity\_tier}. Generate the JSON object as specified in the system
prompt. Remember: L2 must genuinely depend on \texttt{L1\_chosen}; Category B
literals go in the initial question; Category A literals go in L1/L2; both L1 and
L2 must introduce new information; every part of the SQL must be grounded; and
both alternatives must use only the provided schema.

\vspace{0.35em}
\noindent Output protocol.

\noindent
Output only a JSON object containing
\texttt{initial\_ambiguous\_question}, \texttt{L1\_question},
\texttt{L1\_options}, \texttt{L1\_chosen},
\texttt{L1\_alternative\_sql\_impact}, \texttt{L2\_question},
\texttt{L2\_options}, \texttt{L2\_chosen},
\texttt{L2\_alternative\_sql\_impact},
\texttt{L1\_user\_answer\_utterance},
\texttt{L2\_user\_answer\_utterance}, and
\texttt{chain\_rationale}; no markdown fences, reasoning, prose, or commentary.

\end{tcolorbox}
\caption{Compressed version of the prompt used to generate chain ambiguity
samples. The full prompt will be released with the code upon publication.}
\label{fig:chain-prompt}
\end{figure*}

\begin{figure*}[t]
\centering
\begin{tcolorbox}[
  width=1.03\textwidth,
  enhanced,
  colback=white,
  colframe=black,
  boxrule=0.5pt,
  arc=0.8mm,
  left=1.6mm,
  right=1.6mm,
  top=1.3mm,
  bottom=1.3mm
]
\small
\setlength{\parskip}{0.32em}

\noindent System prompt.

\noindent
You are a data annotation specialist constructing INTENT DRIFT samples for a
multi-turn conversational Text-to-SQL benchmark.

\noindent
Core concept: Intent drift is a non-monotonic change in a user's query intent
during conversation. Formally, let $S_{\mathrm{pre}}$ be the set of atomic
commitments encoded in the pre-drift SQL, and $S_{\mathrm{post}}$ be the set of
atomic commitments encoded in the post-drift SQL. A valid drift requires
$S_{\mathrm{pre}} \nsubseteq S_{\mathrm{post}}$, i.e., at least one prior
commitment must be retracted, replaced, or removed. This distinguishes drift from
cooperative follow-up, which is monotonic.

\vspace{0.35em}
\noindent Hard rules.

\begin{enumerate}[leftmargin=1.8em,itemsep=0.22em,topsep=0.15em,label=R\arabic*.]
    \item Non-monotonicity is mandatory.
    The post-drift SQL must retract, replace, or remove at least one commitment
    from the pre-drift SQL. Pure additions, such as appending new predicates while
    keeping all old ones, are not drift.

    \item Drift subtype constraints are strict.
    The assigned drift subtype is one of three. For value-level,
    replace exactly one literal value in a WHERE predicate; SELECT, predicate
    count, and predicate structure must remain identical. For
    constraint-level, delete one or more top-level WHERE predicates;
    SELECT must remain identical, and no new predicates or value changes are
    allowed. For projection-level, change the SELECT projection, including
    columns, aggregations, ORDER BY, or LIMIT; this SELECT shift must be the
    dominant change.

    \item Executability.
    The post-drift SQL must be executable on the given schema, use only existing
    tables and columns, and follow SQLite dialect.

    \item Non-empty result.
    The post-drift SQL must return a non-empty result set.

    \item Different result.
    The post-drift SQL must produce different results from the pre-drift SQL. If
    both return the same rows, the sample fails.

    \item Natural drift utterance.
    \texttt{drift\_utterance} must be natural first-person speech, not a
    mechanical instruction. It should use colloquial transition words such as
    ``wait'', ``actually'', ``forget that'', ``on second thought'', ``scratch
    that'', or ``let me change that''. Length: 1--3 sentences.

    \item Projection-level fine-grained subtype.
    For projection-level only, declare \texttt{intent\_subtype} as either
    \texttt{aspect\_change} or \texttt{target\_change}. In
    \texttt{aspect\_change}, the core entities remain the same, while the
    presentation viewpoint shifts. In \texttt{target\_change}, the core entity,
    grouping key, FROM tables, or WHERE column set changes.
\end{enumerate}

\vspace{0.35em}
\noindent Drift-subtype-specific instructions.

\noindent
\textbf{Value-level:} replace exactly one literal value in a WHERE predicate,
such as changing \texttt{year = 2021} to \texttt{year = 2022}. Do not change
SELECT, GROUP BY, ORDER BY, HAVING, LIMIT, JOIN structure, predicate count,
operator, or filtered column. The new value must be plausible and must produce a
different, non-empty result set. The utterance should explicitly signal which
value is being changed.

\noindent
\textbf{Constraint-level:} delete one or more top-level AND-joined WHERE
predicates. Do not change SELECT, GROUP BY, ORDER BY, HAVING, LIMIT, JOIN
structure, or retained predicate values. Do not remove sub-components of OR
expressions, and do not remove predicates inside nested subqueries. Avoid peeling
off predicates that act as single-row structural anchors; the removal should
broaden an otherwise stable query.

\noindent
\textbf{Projection-level:} change the SELECT projection in a substantive way,
such as changing the returned columns, aggregation, GROUP BY, ORDER BY, or
LIMIT. Declare \texttt{intent\_subtype}. Use \texttt{aspect\_change} when FROM
tables, WHERE column set, and GROUP BY keys remain identical and only the
presentation viewpoint shifts. Use \texttt{target\_change} when at least one of
FROM tables, GROUP BY keys, or WHERE column set differs.

\vspace{0.35em}
\noindent User prompt template.

\noindent
Construct a DRIFT INJECTION sample from a BIRD entry containing
\texttt{db\_id}, \texttt{schema\_ddl}, \texttt{sample\_rows},
\texttt{bird\_question}, \texttt{bird\_evidence}, \texttt{bird\_sql},
\texttt{complexity\_tier}, and \texttt{assigned\_drift\_subtype}. The original
question is the fixed pre-drift user utterance, and \texttt{bird\_sql} is the
pre-drift SQL. Generate only the user's next turn, \texttt{drift\_utterance}, and
the corresponding \texttt{post\_drift\_sql}. The output must satisfy the assigned
drift subtype, executability, non-empty result, different result, and
non-monotonic commitment change.

\vspace{0.35em}
\noindent Output protocol.

\noindent
Output only a JSON object containing
\texttt{drift\_utterance}, \texttt{post\_drift\_sql},
\texttt{drift\_subtype\_claimed}, \texttt{intent\_subtype},
\texttt{pre\_commitments\_narrative},
\texttt{post\_commitments\_narrative}, and \texttt{drift\_rationale}. If
\texttt{drift\_subtype\_claimed} is not projection-level, set
\texttt{intent\_subtype} to null. Return no markdown fences, prose, or commentary
outside the JSON.

\end{tcolorbox}
\caption{Compressed version of the prompt used to generate intent drift
samples. The full prompt will be released with the code upon publication.}
\label{fig:drift-prompt}
\end{figure*}

\begin{figure*}[t]
\centering
\begin{tcolorbox}[
  width=1.03\textwidth,
  enhanced,
  colback=white,
  colframe=black,
  boxrule=0.5pt,
  arc=0.8mm,
  left=1.6mm,
  right=1.6mm,
  top=1.3mm,
  bottom=1.3mm
]
\small
\setstretch{1.04}
\setlength{\parskip}{0.32em}

\noindent System prompt.

\noindent
You are a text-to-SQL agent. Your task is to understand a user's
natural-language data question and produce a correct SQL query for the
database described below.

\vspace{0.35em}
\noindent Database schema.

\noindent
\texttt{\{schema\_ddl\}}

\vspace{0.35em}
\noindent Evidence / domain context.

\noindent
\texttt{\{evidence\}}. Evidence may be empty for some tasks; if so, rely on
schema and dialogue alone.

\vspace{0.35em}
\noindent When to ask vs.\ when to answer.

\noindent
Carefully read the user's request. If anything is ambiguous or
under-specified, ask for clarification rather than making assumptions. When
sufficient information is available, produce the SQL query.

\vspace{0.35em}
\noindent Clarification rules.

\begin{enumerate}[leftmargin=1.8em,itemsep=0.22em,topsep=0.15em,label=R\arabic*.]
    \item Ask at most one clarification question per turn. Focus each
    question on a single ambiguity.

    \item When asking, offer 2--3 discrete options where appropriate to make
    it easy for the user to choose.

    \item Do not ask about information already determinable from the schema
    or evidence above.

    \item Do not ask the user to confirm specific table names, column
    names, or stored values: these are database-internal facts the user
    does not know.
\end{enumerate}

\vspace{0.35em}
\noindent Handling intent change.

\noindent
If the user changes their request mid-dialogue (e.g., ``actually, instead of
X, give me Y''), respect the new request and produce the updated SQL based
on the latest user intent. Do not revert to the original request.

\vspace{0.35em}
\noindent SQL output rules.

\begin{enumerate}[leftmargin=1.8em,itemsep=0.22em,topsep=0.15em,label=S\arabic*.]
    \item When producing SQL, the content must contain only the raw SQL
    query: no explanation, no markdown code fences, no markdown link
    syntax such as \texttt{[col](url)}.

    \item Produce valid SQLite-compatible SQL unless the schema implies
    another dialect.

    \item Use the exact column and table names as they appear in the
    schema, preserving backticks, quotes, and case as shown.
\end{enumerate}

\vspace{0.35em}
\noindent Output protocol.

\noindent
Respond with a single JSON object and nothing else:
\texttt{\{"type": "ask", "content": <clarification question>\}} or
\texttt{\{"type": "sql", "content": <complete SQL query>\}}.

\end{tcolorbox}
\caption{Agent system prompt template used uniformly across all 12 evaluated
models. \texttt{\{schema\_ddl\}} and \texttt{\{evidence\}} are populated
per sample.}
\label{fig:agent-prompt}
\end{figure*}
\end{document}